\documentclass[preprint]{imsart}

\usepackage[pdftex]{graphicx}
\usepackage{subfig}
\usepackage{amsmath,amssymb,amsfonts,mathptmx} 
\usepackage[hyphens]{url}
\usepackage{color}
\usepackage{colortbl}
\usepackage[table]{xcolor}
\usepackage{multirow}
\usepackage{hyperref}
\usepackage{natbib}

\usepackage{threeparttable}
\usepackage{booktabs}
\usepackage{tabularx}
\newcolumntype{C}{>{\centering\arraybackslash}X}

\definecolor{Blue}{rgb}{0.2,0.2,0.9}
\definecolor{rojo}{rgb}{0.843,0.125,0.125}

\definecolor{actionAcolor}{rgb}{1,0.7,0.7}
\definecolor{actionBcolor}{rgb}{0.6,1,0.6}
\newcommand{\actionA}[1]{\cellcolor{actionAcolor} {#1}}
\newcommand{\actionB}[1]{\cellcolor{actionBcolor} {#1}}

\definecolor{actionChangeColor}{rgb}{1,1,0.7}
\definecolor{actionNotChangeColor}{rgb}{0.6,1,1}

\graphicspath{{figs/}}
\graphicspath{{./figures/}}

\newcommand{\Real}{{\mathbb R}}

\newcommand{\vect}[1]{\boldsymbol{\mathbf{#1}}}

\newcommand{\Dict}{\mathbf{D}}

\newcommand{\Coeff}{\mathbf{A}}

\newcommand{\data}{\mathbf{x}}
\newcommand{\Data}{\mathbf{X}}

\newcommand{\lassoenergy}{\mathcal{R}}

\newcommand{\Evol}{\mathbf{E}}
\newcommand{\evol}{E}

\begin{document}

%
%
%
%

\begin{frontmatter}

\title{Are You Imitating Me? \\ 
Unsupervised Sparse Modeling for Group Activity Analysis from a Single Video\thanksref{c1}}
\runtitle{Are You Imitating Me?}

\begin{aug}
  \author{\fnms{Zhongwei}  \snm{Tang}\thanksref{t1,c2}\ead[label=e1]{zhongwei.tang@duke.edu}},
  \author{\fnms{Alexey} \snm{Castrodad}\thanksref{t2}\ead[label=e2]{castr103@umn.edu}}
  \author{\fnms{Mariano}  \snm{Tepper}\thanksref{t1,c2}\ead[label=e3]{mariano.tepper@duke.edu}}
  \and
  \author{\fnms{Guillermo}  \snm{Sapiro}\thanksref{t1,c2}\ead[label=e4]{guillermo.sapiro@duke.edu}}

  \thankstext{c1}{The authors acknowledge partial support from DARPA, ONR, NSSEFF, ARO, NGA, and NSF.}

  \thankstext{c2}{This work was partially done while the authors were with the Department of Electrical and Computer Engineering, University of Minnesota.}

  \runauthor{Z. TANG et al.}

  \affiliation{}

  \address{\thanksmark{t1}Department of Electrical and Computer Engineering, Duke University, USA.\\ 
          \printead{e1,e3,e4}}

  \address{\thanksmark{t2}Department of Electrical and Computer Engineering, University of Minnesota, USA.\\
          \printead{e2}}

\end{aug}

\begin{abstract}
A framework for unsupervised group activity analysis from a single video is here presented. Our working hypothesis is that human actions lie on a union of low-dimensional subspaces, and thus can be efficiently modeled as sparse linear combinations of atoms from a learned dictionary representing the action's primitives. Contrary to prior art, and with the primary goal of spatio-temporal action grouping, in this work only one single video segment is available for both unsupervised learning and analysis without any prior training information. 
After extracting simple features at a single spatio-temporal scale, we learn a dictionary for each individual in the video during each short time lapse. 
These dictionaries allow us to compare the individuals' actions by  producing an affinity matrix which contains sufficient discriminative information about the actions in the scene leading to grouping with simple and efficient tools. 
With diverse publicly available real videos, we demonstrate the effectiveness of the proposed framework and 
its robustness to cluttered backgrounds, changes of human appearance, and action variability.
\end{abstract}

\end{frontmatter}

\section{Introduction}\label{sec:intro}
The need for automatic and semi-automatic processing tools for video analysis is constantly increasing. This is mostly due to the acquisition of large volumes of data that need to be analyzed by a much limited human intervention.  In recent years, significant research efforts have been dedicated to tackle this problem. In this work, we focus on the analysis of human actions, and in particular the spatio-temporal grouping of activities.

Our understanding of human actions and interactions makes us capable of identifying and characterizing these on relative short time intervals and in an almost effortless fashion.  Ideally, we would like to teach a computer system to do exactly this. However, there are challenges that exacerbate the problem, many of which come from the electro-optical system acquiring the data (e.g., noise, jitter, scale variations, illumination changes, and motion blur), but mostly from the inherent complexity and variability of human actions (e.g., shared body movements between actions, periodicity/aperiodicity of body movements, global properties such as velocity, and local properties such as joint dynamics). 

With the availability of large amounts of training data, the above challenges are alleviated to some extent. This is at the foundation of many classification methods that rely on the redundancy of these large datasets, and on the generalization properties of modern machine learning techniques, to properly model human actions. In supervised human action classification, a template model for each class is learned from large labeled datasets. Then, unlabeled actions are classified accordingly to the class model that best represents them.  In this work, we focus on a very different problem, that is, no labeling is available and all data has to be extracted from a single video.\footnote{Even if the video is long, we divide it into short-time intervals to alleviate the action mixing problem. During each short time interval, we have limited data available, and labels are never provided.} A natural question to ask here is \textit{what can we do when only a single unlabeled video is available?} Given such few data, and no {\it a priori} information about the nature of the actions, what we are interested in this work is in human action grouping instead of action recognition. 

Consider for example a camera observing a group of people waiting and moving in line in an airport security checkpoint.  We would like to automatically identify the individuals performing anomalous (out of the norm) actions.  We do not necessarily know what is the normal action nor the anomalous one, but are interested in knowing when a ``different from the group" action is occurring on a given time lapse, and in being able to locate the corresponding individual (in space and time).  Situations like this not only occur in surveillance applications, but also in psychological studies (i.e., determining outlier autistic behavior in a children's classroom, or identifying group leaders and followers), and in the sports and entertainment industry (e.g., identifying the offensive and defensive teams, or identifying the lead singer in a concert).  

We focus on modeling the general dynamics of individual actions in a {\it single} scene, with no {\it a priori} knowledge about the actual identity of these actions nor about the dynamics themselves.  We propose an intuitive unsupervised action analysis framework based on sparse modeling for space-time analysis of motion imagery. The underlying idea we propose is that the activity dictionary learned for a given individual is also valid for representing the same activity of other individuals, and not for those performing different ones, nor for him/her-self after changing activity. We make the following main contributions:

\begin{itemize}

\item {\bf Unsupervised action analysis:} We extend the modeling of human actions in a relatively unexplored area of interest. That is, we analyze unknown actions from a group of individuals during consecutive short-time intervals, allowing for action-based video summarization from a single video source.

\item {\bf Solid performance using a simple feature:}  We use a simple feature descriptor based on absolute temporal gradients, which, in our setting, outperforms more sophisticated alternatives.

\item {\bf Sparse modeling provides sufficient discriminative information:} We demonstrate that the proposed sparse modeling framework efficiently separates different actions and is robust to visual appearance even when using a single basic feature for characterization and simple classification rules.

\item {\bf Works on diverse data:} We provide a simple working framework for studying the dynamics of group activities, automatically detecting common actions, changes of a person's action, and different activities within a group of individuals, and test it on diverse data related to multiple applications.
\end{itemize}

The remainder of the paper is structured as follows.
In Section~\ref{sec:review}, we provide an overview of recently proposed methods for supervised and unsupervised action classification. Then, in Section~\ref{sec:method}, we give a detailed description of the proposed modeling and classification framework.  We demonstrate the pertinence of our framework in Section \ref{sec:experiments} with action grouping experiments in diverse (both in duration and content) videos.
Finally, we provide concluding remarks and directions for future work in Section~\ref{sec:conclusion}.

\section{Background and model overview}\label{sec:review}




In this section, we review recent techniques for human action classification which are related to the present work. We  focus on feature extraction and modeling, and cover both supervised and unsupervised scenarios.

\subsection{Features for action classification} \label{subsec:features}

Most of the recently proposed schemes for action classification in motion imagery are feature-based.  In general, one could split the feature extraction process into an interest point detection phase and a descriptor encoding phase. Interest point detection consists in finding spatio-temporal locations across the video where a pre-defined response function achieves local extrema, e.g., high spatio-temporal variations. Popular detectors are the Spatio-temporal Interest Point detector (STIP) \citep{Laptev05}, Cuboids \citep{dollar05}, and Hessian~\citep{willems08}.  Feature descriptors include the Cuboid feature~\citep{dollar05}, Histogram of 3D Oriented Gradients (HOG3D)~\citep{klaser08}, the combination of HOG and Histograms of Optical Flow (HOF) ~\citep{laptev08}, Local Motion Patterns (LMP)~\citep{Guha2012}, and  Extended Speeded Up Robust Features (ESURF)~\citep{willems08}, most of which are spatio-temporal extensions to techniques designed for still images.

In practice, it is unclear which interest point detector and feature combination is the most appropriate for modeling human actions.
\citet{wang09} performed an exhaustive comparison of different detector/feature combinations on several datasets. Although individual performance depends on the dataset, dense sampling (no interest point detection) combined with HOG/HOF features seems a good choice in realistic video settings, since it captures context from the scene background. \citet{shao10} performed a similar evaluation using a less realistic dataset (the KTH action dataset\footnote{\url{http://www.nada.kth.se/cvap/actions/}}) and concluded that the Cuboid detector combined with the Local Binary Pattern on Three Orthogonal Planes (LBP-TOP) descriptor~\citep{lbp,zhao07} gives the best performance in terms of classification accuracy. 

Apart from these localized, low-level descriptors, there has been research focused on the design of more global, high-level descriptors, which encode more semantic information.  For example, \citet{actionbank} proposed to represent the video as the collected output from a set of action detectors sampled in semantic and in viewpoint space. Then, these responses are max-pooled and concatenated to obtain semantic video representations. \citet{actionshape} proposed to model human actions as 3D shapes on a space-time volume. Then, the solution of the Poisson equation is used to extract different types of feature descriptors to discriminate between actions. \citet{Sketch2005} proposed to use a similar space-time volume by calculating the point correspondences between consecutive frames, and action representative features were computed by differential geometry analysis of the space-time volume's surface. Other types of high-level features are the joint-keyed trajectories or human pose, which have been used for example by \citet{ramanan2003} and \citet{Shah2007}. \citet{kliper12} proposed to encode entire video clips as single vectors by using Motion Interchange Patterns (MIP), which encode sequences by comparing patches  between three consecutive frames and applying a series of processes for background suppression and video stabilization.


Finally, it is clear that the choice of detectors and descriptors and their respective performances highly depend on the testing scenarios, including acquisition properties, dataset physical settings, and the modeling techniques.  Most of these features work well in the context for which they were proposed, and changing the context might adversely affect their performance. Let us emphasize that feature design is not our main goal in this paper. We next describe the feature extraction scheme  used throughout this work, which, although very simple, works very well in all our scenario, hence highlighting the advantages of the very simple proposed model.

\begin{figure}
	\centering
	\includegraphics[width=\textwidth]{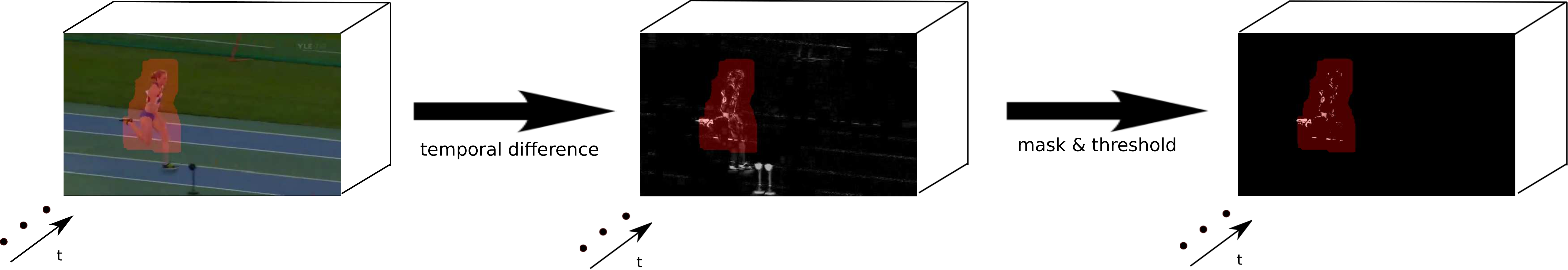}
	\caption{Scheme of the interest point extraction method. For extracting the interest points for a given individual, we keep the points (1) whose temporal gradient exceeds a given threshold and (2) that lie inside the individual's (dilated) segmentation mask. The contrast of the temporal gradient is enhanced for improved visualization. }
	\label{fig:feature_chain}
\end{figure}

\paragraph{\textbf{The proposed feature.}}
In order to properly capture the general spatio-temporal characteristics of actions, it is always desirable to have a large number of training samples.  We aim at characterizing actions from scarce data and, under these conditions, we are able to properly model the actions using a simple feature (the overall scheme is illustrated in Fig.~\ref{fig:feature_chain}). 
We start by tracking and segmenting \citep{Papadakis2011} the individuals whose actions we analyze. This segmentation masks allow us to focus mostly on the individuals while disregarding (most of) the background. We then set a simple interest point detector based on the absolute temporal gradient. For each individual, the points where the absolute temporal gradient is large enough (i.e., it exceeds a pre-defined threshold) become interest points for training and modeling. The feature is also very simple: it consists of a 3D (space and time) absolute temporal gradient patch around each interest point.
As we will illustrate in Section \ref{sec:space_exp}, this combination works better than some more sophisticated alternatives in the literature.
 
\subsection{Modeling actions for classification} \label{subsec:modeling}

Once the feature samples are extracted from the video, an appropriate model is necessary to obtain valid and discriminative action representations. Bag-of-words is one of the most widely used models~\citep{wang09,shao10}.  It basically consists of applying K-means clustering to the data to find K centroids, i.e., visual words, that are representative of all the training samples.  Then, a video is represented as a histogram of visual word occurrences, assigning one of the centroids to each of the extracted features in the video using (most often) Euclidean distance.   These K centroids are found using a randomly selected subset of features coming from all the training data. Then, a classifier like a Support Vector Machine (SVM) can be trained from these histograms.

Recently, sparse modeling has proven to be very successful in signal and image processing applications, especially after highly efficient optimization methods and supporting theoretical results emerged. The advantage of sparse modeling over other dictionaries fixed a priori, like Fourier or wavelet basis, is that it learns the dictionary from the data themselves and can thus represent them more efficiently. Sparse modeling allows to represent each sample as a linear combination of a few atoms in the dictionary (a subspace), as opposed to the single ``atom'' representation in standard bag-of-words. This means that sparse modeling is more flexible but still keeps a rich representation power, in contrast to bag-of-words representations, which are hard quantizations of the data samples, regardless of their distance from the closest centroid.
Sparse modeling has been adapted to classification tasks like face recognition \citep{wright08} (without dictionary learning), digit and texture classification~\citep{mairal08}, hyperspectral imaging~\citep{castrodad11,charles11}, and motion imagery \citep{cadieu08,dean09,guo10,taylor10}, among numerous other applications (note that techniques based on sparse modeling have also performed very well in the PASCAL competition). 

Several sparse modeling approaches have been recently proposed for action classification tasks as well. 
\citet{Guha2012} used learned dictionaries in three ways:
individual dictionaries (one per action), a global (shared) dictionary, and a concatenated dictionary.
Individual dictionaries are separately learned for each class of actions and unlabeled actions are assigned to the class whose dictionary gives the minimum reconstruction error.  The concatenated dictionary is formed by concatenating all the individual dictionaries, and unlabeled actions are assigned to the class whose corresponding subdictionary contributes the most to the reconstruction.  To create the (shared) dictionary, a single common and unstructured dictionary is learned using all training feature data from every class. The dictionary coding coefficients of training actions are used to train a multi-class SVM.  A shared (global) dictionary was also proposed by \citet{dean09}, where a dictionary is learned in a recursive manner by first extracting high response values coming from the Cuboids detector, and then using the resulting sparse codes as the feature descriptors (PCA is optionally applied). Then, as often done for classification, the method uses a bag-of-features approach for representing the videos and a nonlinear $\chi^2$-SVM for classification. \citet{guo10} built a dictionary using vectorized log-covariance matrices of $12$ hand-crafted features (mostly derived from optical flow) obtained from entire labeled videos.  Then, the vectorized log-covariance matrix coming from an unlabeled video is represented with this dictionary using $\ell_1$-minimization, and the video is classified by selecting the label associated with those dictionary atoms that yields minimum reconstruction error. 
\citet{castrodad211} propose a two-level sparse modeling scheme, in order to capture shared movements from different actions, achieving highly accurate classifications.  Sparse modeling has also been applied to abnormal event detection, which can be considered as a binary classification problem. \citet{abnormalevent} proposed a model to detect abnormal events by means of high reconstruction errors obtained by encoding using a ``normal events'' dictionary, which is constructed using a collection of local spatio-temporal patches from ``normal events'' sequences. 

More recently, human action models have been extended to account for human interactions and group activities. \citet{khamis12} introduced a model to classify nearby human activities by enforcing homogeneity on both the identity and the scene context on a frame by frame basis. \citet{todorovic12} proposed to detect and localize individual, structured, and collective human activities (segmented as foreground) by using Kronecker (power) operations on learned activity graphs, and then classify these based on permutation-based graph matching. \citet{fu12} proposed a model to label group (social) activities using audio and video by learning latent variable spaces of user defined, class-conditional, and background attributes. \citet{choi12} proposed to track and estimate collective human activities by modeling label information at several levels of a hierarchy of activities going from individual to collective, and encoding their respective correlations. Our work is similar to these in the sense that we seek to analyze group activities by 
exploiting the correlations of individual's actions.  However, all of the above mentioned schemes require a large amount of labeled training data, which are not available for single video analysis.  For this reason, we now turn the attention to unsupervised approaches.

\subsection{Unsupervised setting} \label{subsec:unsupervised}

In multi-class supervised classification, labeled training samples from different classes are required, and for anomalous events detection, ``normal'' training samples are needed. The majority of the publicly available data benchmarks for human action classification usually contain only one person and one type of action per video, and are usually accompained by tracking bounding boxes \citep{Hassner2012}. This is different from our testing scenario, where only a single video (segment) is available, containing more than one person, without other prior annotations.

Several works addressing unsupervised human action classification have been proposed. \citet{niebles08} used probabilistic Latent Semantic Analysis (pLSA) and Latent Semantic Analysis (LSA) to first learn different classes of actions present in a collection of unlabeled videos through bag-of-words, and then apply the learned model to perform action categorization in new videos. This method is improved 
by using spatio-temporal correlograms to encode long range temporal information into the local features \citep{savarese08}. 
\citet{willem09} proposed a bag-of-words approach using Term Frequency-Inverse Document Frequency features and a data-stream clustering procedure. A spatio-temporal link analysis technique combined with spectral clustering to learn and classify the classes was proposed by \citet{liu09}. 
All of these methods employ a bag-of-words approach with sophisticated features and require training data to learn the action representations, while we only work on one single video segment, and a much simpler feature.
Our work also departs from correlation-based video segmentation. These methods usually correlate a 
sample video clip with a long video to find the similar segments in the target video \citep{Irani2007}, while our work treats all the actions in one video equally and automatically find the groups of the same action.
The work presented here shares a similar (but broader) goal as Zelnik-Manor and Irani's \citeyearpar{Irani2005}.
Their unsupervised action grouping technique works for a single video containing one individual, comparing the histograms of 3D gradients computed throughout each short-time intervals. We consider a more general setting in which the video length ranges between one second to several minutes, and contains more than one individual, with individuals performing one or more actions and not necessarily the same action all the time. 

Bearing these differences in mind, we now proceed to describe the proposed model in detail. 

\section{Unsupervised modeling of human actions}
\label{sec:method}

In this work, we assume there are $P \geq 1$ individuals performing simultaneous actions in the video.
We first use an algorithm based on graph-cuts \citep{Papadakis2011} to coarsely track and segment the individuals. These tracked segmentations will be used as masks from which features for each individual will be extracted. We later show that these coarse masking procedure is sufficient for reliably grouping actions with our method.

We first extract spatio-temporal patches from the absolute temporal gradient image, around points which exceeds a pre-defined temporal gradient threshold $\eta$. These $m$-dimensional spatio-temporal patches from the $j$-th person are the data used to train the corresponding dictionary $\Dict^j, j = 1, 2, \cdots, P$.  Let us denote by $n_j$ the number of extracted patches from the $j-th$ individual.
More formally, we aim at learning a dictionary $\Dict^j \in \Real^{m \times k_j}$ such that a training set of patches $\Data^j = [ \data_1, \dots, \data_{n_j} ] \in \Real^{m \times n_j}$ can be well represented by linearly combining a few of the basis vectors formed by the columns of $\Dict^j$, that is $\Data^j \approx \Dict^j \Coeff^j$. Each column of the matrix $\Coeff^j \in \Real^{k_j \times n_j}$ is the sparse code corresponding to the patch from $\Data^j$. In this work we impose an additional nonnegativity constraint on the entries of $\Dict^j$ and $\Coeff^j$. This problem can then be casted as the optimization
\begin{equation}    \label{eq:sparseModeling}
   \min_{(\Dict^j , \Coeff^j) \succeq 0} \ \frac{1}{2} \| \Data^j - \Dict^j \Coeff^j \|_F^2 + \lambda \| \Coeff^j \|_{1,1} ,
\end{equation}
where $\succeq$ denotes the element-wise inequality, $\lambda$ is a positive constant controlling the trade-off between reconstruction error and sparsity (numerous techniques exist for setting this constant \cite[e.g.,][]{tibshirani94}), $\|\bullet\|_{1,1}$ denotes the $\ell_1$ norm of a matrix, that is, the sum of its coefficients, and $\|\bullet\|_F$ denotes the Frobenius norm. 
Since Equation~(\ref{eq:sparseModeling}) is convex with respect to the variables $\Coeff^j$ when $\Dict^j $ is fixed and vice versa, it is commonly solved by alternatively fixing one and minimizing over the other.\footnote{Recent developments, e.g., \cite{Ramadge11,LeCun10,mairal10,Bronstein2012}, have shown how to perform dictionary learning and sparse coding very fast, rendering the proposed framework very efficient.}

\begin{figure}[ht]
\centering
\includegraphics[width=0.8\textwidth]{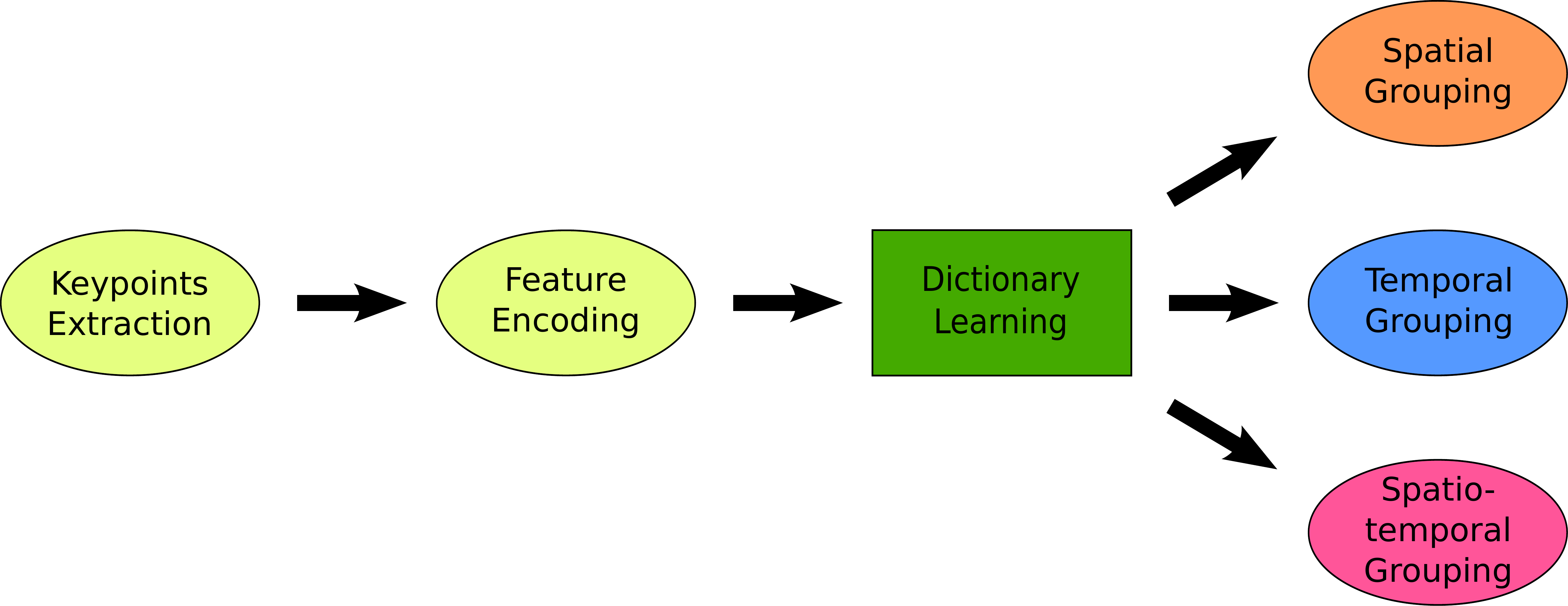} \\
\caption{Algorithmic pipeline of the proposed method. The first three stages (keypoints extraction, feature extraction and dictionary learning) are common to all presented analysis tools, while specific techniques at the pipeline's end help answer different action-grouping questions.
Although we propose tools for solving all the different stages in this pipeline, the core contribution of this work is in the modeling of actions via dictionary learning (the corresponding stage is denoted by a rectangle). This  
allows to use very simple techniques in the previous stages and much flexibility in the subsequent ones.}
\label{fig:pipeline}
\end{figure}

We will show next how to use these learned dictionaries for comparing simultaneously performed actions on a single time interval (Section~\ref{sec:space}), and to detect action changes of the individuals along the temporal direction (Section \ref{sec:time}), with 
a special case when $P = 2$ in Section \ref{sec:special}. Finally, a spatio-temporal joint grouping for a long video is presented in Section 
\ref{sec:joint_grouping}. The algorithm's pipeline is outlined in Fig.~\ref{fig:pipeline}. 

\subsection{Comparing simultaneously performed actions} \label{sec:space}

On a standard supervised classification scenario, subdictionaries $\Dict^j$  are learned for each human action, and are concatenated together to form a global dictionary. Then, new unlabeled data from human actions are represented by this global dictionary and are classified into the class where the corresponding subdictionary plays the most significant role in the reconstruction \citep{castrodad211}. In our case, we do not have labeled data for learning and classification. During reconstruction, each person will obviously tend to prefer its own subdictionary from the global dictionary (since the subdictionary is learned from the very same data), consequently inducing poor discrimination power.  To handle this difficulty, a ``leave-one-out'' strategy is thus proposed: each individual action $j$ is represented in a global dictionary that excludes its corresponding subdictionary $\Dict^j$. 
 
Let us assume that for each person $j \in [1, P]$, we have learned a dictionary $\Dict^j \in \Real^{m \times k_j}$ using the patches  
$\Data^j$. We concatenate the $P-1$ dictionaries $\{ \Dict^i \}_{i = 1 \dots P, i \neq j}$ (that is, without including $\Dict^j$), to build the dictionary $\overline{\Dict^j} \in \Real^{m \times k}$, $k = \sum_ {i=1, i \neq j}^{P} k_i$. To test the similarity between the action performed by the $j$-th person and those performed by the rest of the group, we solve
\begin{equation}
    \min_{\overline{\Coeff^j} \succeq 0} \ \frac{1}{2} \| \Data^j - \overline{\Dict^j} \ \overline{\Coeff^j} \|_F^2 + \lambda \| \overline{\Coeff^j} \|_{1,1} \ .
    \label{eq:sparseCoding}
\end{equation}
The computed sparse-codes matrix $\overline{\Coeff^j}$ is the concatenation of the sparse codes blocks $\{ \overline{\Coeff^{j, i}} \}_{i = 1 \dots P, i \neq j}$ such that
\begin{align*}
    \overline{\Dict^j} \ \overline{\Coeff^j} & = 
    \begin{bmatrix}
        \Dict^1, \dots, \Dict^{j-1}, \Dict^{j+1}, \dots,  \Dict^P
    \end{bmatrix}
    \begin{bmatrix}
        \overline{\Coeff^{j, 1}} ^T
        \dots 
        \overline{\Coeff^{j, j-1}} ^T, 
        \overline{\Coeff^{j, j+1}} ^T
        \dots 
        \overline{\Coeff^{j, P}}^T
    \end{bmatrix} ^T\\
    & = \Dict^1 \overline{\Coeff^{j,1}} + \cdots + \Dict^{j-1} \overline{\Coeff^{j,j-1}} + \Dict^{j+1} \overline{\Coeff^{j,j+1}} \cdots + \Dict^P \overline{\Coeff^{j,P}}. 
\end{align*}
We use $\| \overline{\Coeff^{j, i}} \|_{1,1}$ to encode the level of similarity between the action corresponding to the $j$-th person and the action corresponding to the $i$-th person, $\forall i\neq j$. Let us motivate this choice with the following example. If two persons, $j$ and $i$ are performing similar actions and person $i'$ is performing a different action, when trying to represent $\Data^j$ with the dictionary $\overline{\Dict^j}$, a larger $\ell_1$ energy (activation) is expected from the block $ \overline{\Coeff^{j, i}}$ (corresponding to $\Dict^i$) than from that of $ \overline{\Coeff^{j, i'}}$ (corresponding to $\Dict^{i'}$).
We then define the action-similarity matrix $\vect{S} \in \Real^{P \times P}$, whose entries $s_{ij}$ are defined as
\begin{equation}
    s_{ij} = 
    \begin{cases}
    \min \left( \frac{\| \overline{\Coeff^{i,j}} \|_{1,1}}{ \| \overline{\Coeff^i}  \|_{1,1}} \ , \  \frac{\| \overline{\Coeff^{j,i}} \|_{1,1}}{ \| \overline{\Coeff^j}  \|_{1,1}} \right) & \text{if } i \neq j \ , \\
    1 & \text{otherwise} \ . \\
    \end{cases}
    \label{eq:similarityMatrix}
\end{equation}
The minimum is used to enforce reciprocal action similarity, and the normalization ensures that comparisons between all individual actions are fair. 

We then consider the matrix $\vect{S}$ as the affinity matrix of a nonoriented weighted graph $G$. Although numerous techniques can be used to partition $G$, in this work we use a simple approach that proved successful in our experiments (recall that the expected number of persons $P$ is small in a group, in contrast to a crowd, so clustering techniques, which rely on statistical properties of the graph, are neither needed nor appropriate). We simply remove the edges of $G$ that correspond to entries $s_{ij}$ such that $s_{ij} < \tau$, for a given threshold $\tau$. For a properly chosen threshold, this edge removal will cause $G$ to split into several connected components, and we consider each one as a group of persons performing the same action.  In an ideal scenario where all actions are equal, the similarity scores $\| \overline{\Coeff^{j,i}} \|_{1,1}$ ($i = 1,\dots, P, i \neq j$) will also be similar. Since in Equation (\ref{eq:similarityMatrix}) we normalize them, setting the threshold to $1/(P-1)$ seems to be a natural choice. However, in practice, the distribution of these coefficients is not strictly uniform. 
For example,  in a video with four skeletons dancing  in a synchronous fashion (see Fig.~\ref{fig:skeleton_dancing_example}), the similarity scores in the resulting affinity matrix still show slight variations.  
We thus set $\tau = \frac{r}{P-1}$,  
where $r \in [0, 1]$ is a relaxation constant (in our experiments we found that $r=0.9$ was sufficient to cope with this nonuniformity effect).

\subsection{Temporal analysis: Who changed action?} \label{sec:time}

In the previous section, we presented the modeling and grouping scheme for a fixed time interval (a given video segment). The matrix $\vect{S}$ provides sufficient information to determine if there are different actions occurring during an interval, and to determine which individual/s are performing them.\footnote{The same framework can be applied if we have a single individual and just want to know if all the activities he/she is performing in $P>2$ time intervals are the same or not, each time interval taking the place of an ``individual.''}  Suppose that on a given interval $t-1$, all $P$ individuals are performing the same action, then, on the next time interval $t$, the first $P-1$ individuals change action while the $P$-th individual remains doing the same. From the affinity matrix $\vect{S}$ there is no way of determining if the first $P-1$ persons changed while the $P$-th person remained doing the same or vice-versa.  An additional step is thus necessary in order to follow the individuals' action evolution in the group.  A possible solution for this problem at a small additional computational cost is as follows.

Let the minimized energy
\begin{equation}
   \lassoenergy^*(\Data_t^j,\Dict_t^j) = \min_{\Coeff^j \succeq 0} \ \frac{1}{2} \| \Data_t^j- \Dict_t^j \Coeff^j \|_F^2 + \lambda \| \Coeff^j \|_{1,1} \,
\end{equation}
be the $j$-th individual's $\ell_{2,1}$ representation error with his/her own dictionary at time $t$.  Then, we measure the evolution of the reconstruction error per individual as
\begin{equation}
{E}^{j}_{t-1,t}= |(\lassoenergy^*(\Data^j_{t-1},\Dict_t^j) + \lassoenergy^*(\Data^j_t,\Dict^j_{t-1}) - \lassoenergy^*(\Data^j_{t-1},\Dict^j_{t-1}) - \lassoenergy^*(\Data^j_t,\Dict^j_t))|,
\end{equation}
and 
\begin{equation}
\Evol_{t-1,t} = \frac{1}{C} [\evol^{1}_{t-1,t},...,\evol^{P}_{t-1,t}] ,
\label{eq:temporalEnergyVector}
\end{equation}
where $C = \sum_{j=1}^P \evol^{j}_{t-1,t}$ is a normalization constant. 
$\Evol_{t-1,t}$ captures the action changes in a 
per person manner, a value of $\evol^{j}_{t-1,t} / C$ close to $1$ implies that the representation for the $j$-th individual has changed drastically, 
while $0$ implies the individual's action remained exactly the same. 
In the scenario where nobody changes action, all $\evol^{j}_{t-1,t}, \forall j\in [1,P]$ will be similar. 
We can apply a similar threshold $\mu = r/P$ to detect the actions' time changes (note that we now have $P$ persons instead of $P-1$).

\subsection{Special case: $P = 2$} \label{sec:special}
If there are only two individuals in the video ($P=2$), the grouping strategies from sections~\ref{sec:space} and~\ref{sec:time} become ambiguous. The similarity matrix $\vect{S}$ would have all entries equal to $1$ (always one group), and there would be no clear interpretation of the values in $\Evol_{t-1,t}$. Therefore, for this particular case where $P=2$, we define the one time interval measure (at time $t$) as
\begin{equation} \label{eq:space_special_case}
 \evol^{i,j}_t = \max\Bigg(\frac{|\lassoenergy^*(\Data^{i}_t,\Dict^j_t) - \lassoenergy^*(\Data^i_{t},\Dict^i_{t})|}{|\lassoenergy^*(\Data^i_t,\Dict^i_t)|}, \frac{|\lassoenergy^*(\Data^j_{t},\Dict^i_t) - \lassoenergy^*(\Data^j_{t},\Dict^j_{t})|}{|\lassoenergy^*(\Data^j_{t},\Dict^j_{t})|}\Bigg),  
\end{equation}
where $i$ and $j$ denote each subject.  Also, we let the evolution of the reconstruction error from $t-1$ to $t$ of each individual $i \in [1,2]$ be 
\begin{equation} \label{eq:time_special_case}
\evol^{i}_{t-1,t} = \max\Bigg(\frac{|\lassoenergy^*(\Data^{i}_{t-1},\Dict^i_{t}) - \lassoenergy^*(\Data^i_{t-1},\Dict^i_{t-1})|}{|\lassoenergy^*(\Data^i_{t-1},\Dict^i_{t-1})|}, \frac{|\lassoenergy^*(\Data^i_{t},\Dict^i_{t-1}) - \lassoenergy^*(\Data^i_{t},\Dict^i_{t})|}{|\lassoenergy^*(\Data^i_{t},\Dict^i_{t})|}\Bigg). 
\end{equation}
Finally, we use the threshold $\mu = r/2$, where a value greater than $\mu$ implies that the subject is performing a different action.
Note that these formulations do not replace the algorithms in sections~\ref{sec:space} and~\ref{sec:time}, which are more general 
and robust to treat the cases where $P > 2$.

\subsection{Joint spatio-temporal grouping} \label{sec:joint_grouping}

The above discussion holds for a short-time video containing a few individuals. Clustering techniques, which rely on statistical properties of the data, are neither needed nor appropriate to handle only a few data points. Simpler techniques were thus needed.

Now, for long video sequences, action analysis with the previously described simple tools becomes troublesome. Luckily, clustering methods are ideal for this scenario.
Hence, we use them to do joint spatio-temporal action grouping (here, by spatial we mean across different individuals).
We consider
each individual in a given time interval as a separate entity (an individual in two different time intervals is thus considered as two individuals).
Dictionaries are learned for each spatio-temporal individual and an affinity matrix is built by comparing them in a pairwise manner using equations~(\ref{eq:space_special_case}) or (\ref{eq:time_special_case}) (notice that in this setting, the two equations become equivalent). 
We simply apply to this affinity matrix a non-parametric spectral clustering method that automatically decides the number of groups \citep{Perona2004}.


\section{Experimental results}\label{sec:experiments}
In all the reported experiments, we used $n=\min \left( \min_j (n_j), 15,000\right)$ overlapping temporal gradient patches of size $m = 15 \times 15 \times 7$ for learning a dictionary per individual. The tracked segmentation mask for each individual \citep{Papadakis2011} is dilated to ensure that the individual is better covered (sometimes the segmentation is not accurate for the limbs under fast motions). Only the features belonging to the tracked individuals are used for action modeling and dictionary learning. The dictionary size was fixed to $k_j = 32$ for all $j$, which is very small compared to the patch dimension (undercomplete). The duration of the tested video segments (short-time intervals) is one second for action grouping per time interval, and two seconds for temporal analysis. Longer videos (from 5 seconds to 320 seconds) were used for joint spatio-temporal grouping. 
All the tested videos are publicly available and summarized in Table~\ref{table:video_table}.\footnote{We only present a few frames for each video in the figures, please see the supplementary material and mentioned links for the complete videos.}

\begin{table}
\caption{Description summary of the videos used in the experiments.}
\label{table:video_table}

\centering
\begin{threeparttable}[b]
\begin{tabularx}{\textwidth}{llcl}
\toprule
Video & fps & Figures & Description\\
\midrule
Skeleton\tnote{a} &  30 & \ref{fig:skeleton_dancing_example} & Four skeletons dancing in a similar manner.\\ 
Long jump\tnote{b} &  25 & \ref{fig:imgs_running_long_jumping}, \ref{fig:long_jumping_clustering} & Three athletes in a long jump competition.\\
Gym\tnote{c} & 30 & \ref{fig:imgs_punching_dancing} & Three persons in a gym class.\\
Kids\tnote{d} & 30 & \ref{fig:imgs_kids} & Three kids dancing in a TV show.\\
Crossing\tnote{e,f} & 30 & \ref{fig:crossing_waiting_example} & Two pedestrians crossing the street and the other one waiting.\\
Jogging\tnote{g,f} & 30 & \ref{fig:jogging_example} & Six persons jogging in a park.\\
Dancing\tnote{h,f} & 30 & \ref{fig:dancing_example} & Five persons rehearsing a dance act.\\
Singing-dancing\tnote{i} & 30 & \ref{fig:taiwanese_dancing_example} & A singer and four dancers performing in a theater.\\ 
Tango\tnote{j} &  30 & \ref{fig:Tango_dancing_example} & Three couples dancing Tango. \\ 
Mimicking\tnote{k} & 30 & \ref{fig:mimicking} & Gene Kelly in a self-mimicking dancing act.\\ 
Fitness\tnote{l} &   30 &  \ref{fig:action_evolution}, \ref{fig:warmingup_clustering} & Three persons in a fitness class.\\ 
Outdoor\tnote{m} &  30 & \ref{fig:outdoor_clustering} & Action classification video used by \citet{Irani2005}\\ 
\bottomrule 
\end{tabularx}
\begin{scriptsize}
\begin{tablenotes}
\begin{scriptsize}
\item[a] \url{http://www.youtube.com/watch?v=h03QBNVwX8Q}
\item[b] \url{http://www.youtube.com/watch?v=bia-x_linh4}.
\item[c] \url{http://www.openu.ac.il/home/hassner/data/ASLAN/ASLAN.html}
\item[d] \url{http://www.youtube.com/watch?v=N0wPQpB4eMk}
\item[e] \url{http://www.eecs.umich.edu/vision/activity-dataset.html}
\item[f] Bounding boxes surrounding each person are provided, and they were used instead of the tracking/segmentation approach.
\item[g] \url{http://www.eecs.umich.edu/vision/activity-dataset.html}
\item[h] \url{http://www.eecs.umich.edu/vision/activity-dataset.html}
\item[i] \url{http://www.youtube.com/watch?v=R9msiIqkI34}
\item[j] \url{http://www.youtube.com/watch?v=IkFjg7m-jzs}
\item[k] \url{http://www.youtube.com/watch?v=_DC6heLMqJs}
\item[l] \url{http://www.youtube.com/watch?v=BrgPzp0GBcw}
\item[m] \url{http://www.wisdom.weizmann.ac.il/mathusers/vision/VideoAnalysis/Demos/EventDetection/OutdoorClusterFull.mpg}

\end{scriptsize}
\end{tablenotes}
\end{scriptsize}
\end{threeparttable}
\end{table}

We provide a rough running-time estimate, in order to show the efficiency of the proposed framework.
Our non-optimized Matlab code for feature extraction takes less than 10 seconds to process 30 frames of a standard VGA video ($640 \times 480$ pixels), containing five individuals, and about 3 seconds for 30 frames of a  video with lower-resolution ($480 \times 320$ pixels) containing three individuals.
As for the sparse modeling, we used the SPAMS toolbox,\footnote{\url{http://spams-devel.gforge.inria.fr/}} taking approximately 1 second to perform dictionary learning and sparse coding of 15,000 samples. Notice that  code optimization and parallelization would significantly boost the performance, potentially obtaining a real-time process.

\subsection{Action grouping per time interval} \label{sec:space_exp}

\begin{figure}[ht]
\centering
\begin{tabular}{@{\hspace{0pt}}m{.27\textwidth}*{2}{@{\hspace{4pt}}m{.33\textwidth}}@{\hspace{0pt}}}
\begin{minipage}{.27\textwidth}
\centering
\includegraphics[width=\textwidth]{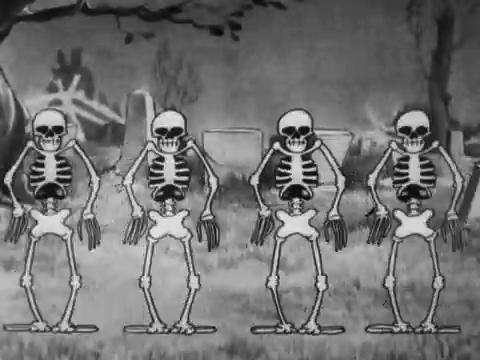}\\
\includegraphics[height=0.115\textheight]{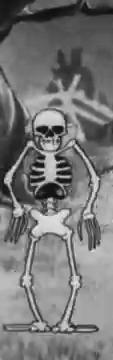} 
\includegraphics[height=0.115\textheight]{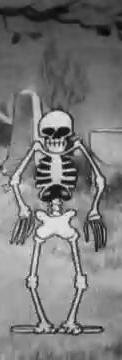} 
\includegraphics[height=0.115\textheight]{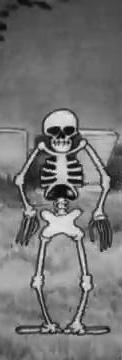} 
\includegraphics[height=0.115\textheight]{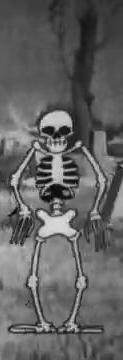} 
\end{minipage} &
\includegraphics[width=.33\textwidth]{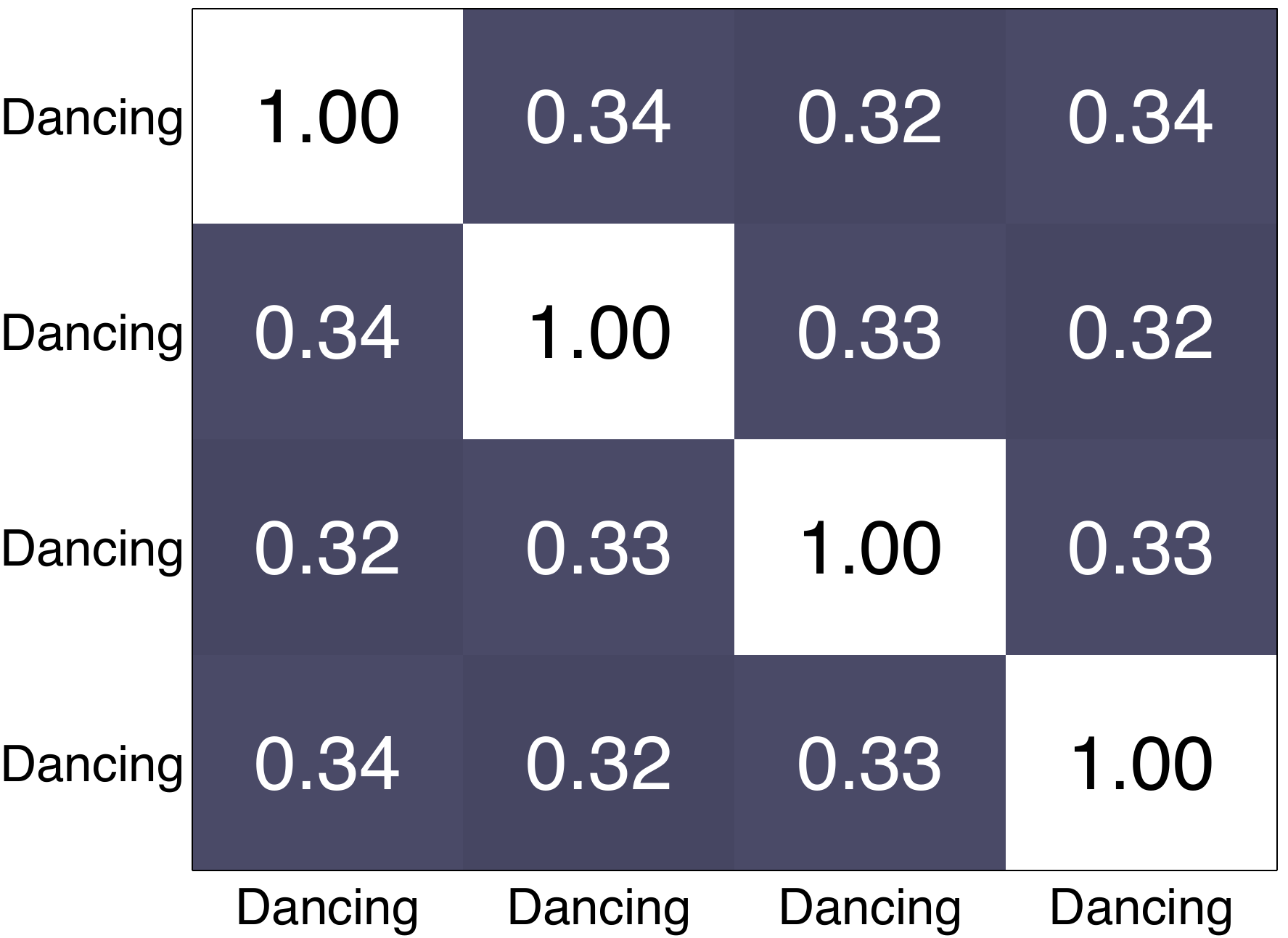} & 
\includegraphics[width=.33\textwidth]{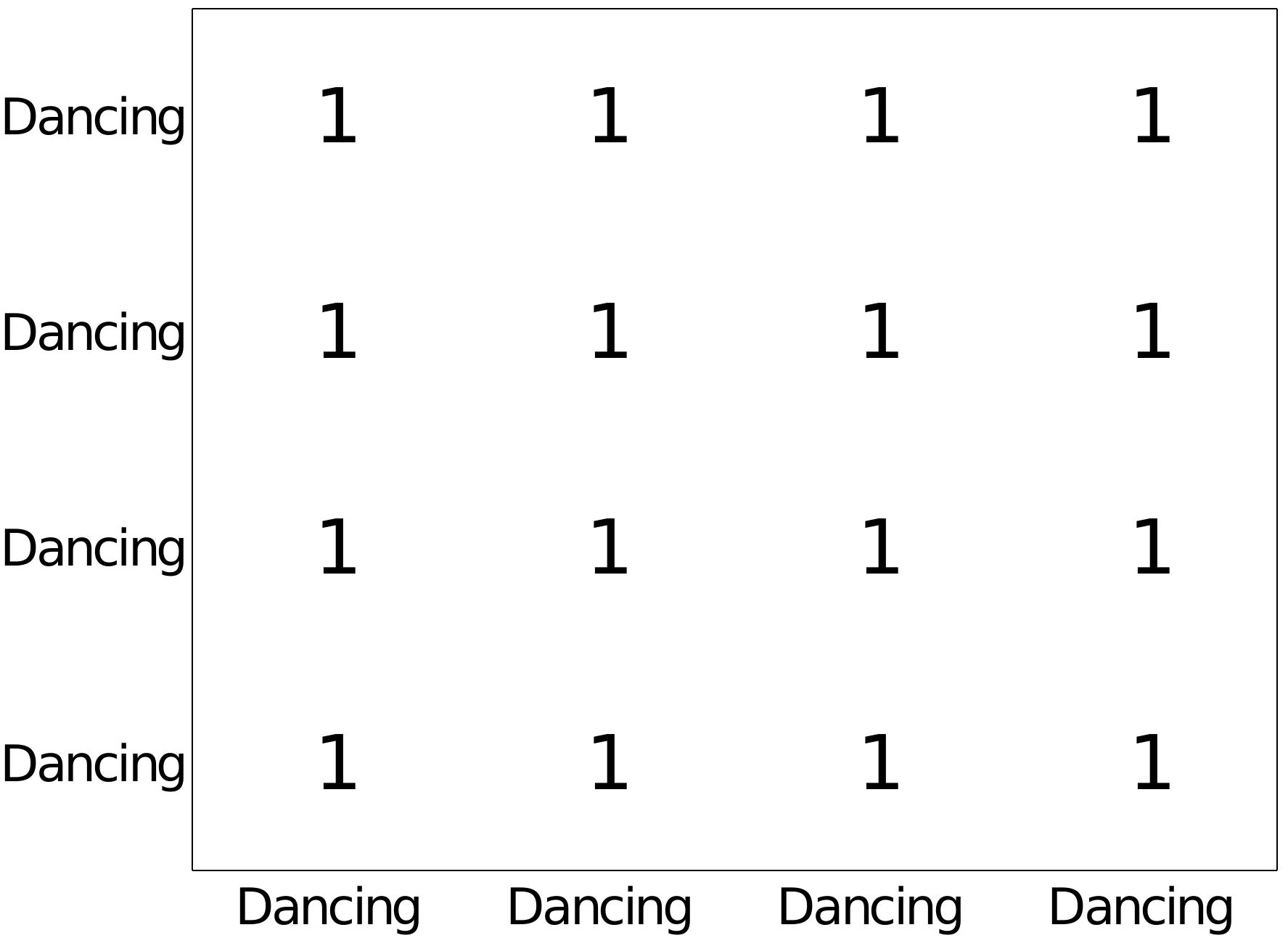} \\
\end{tabular}
\caption{Sample frames from the Skeleton video. \textbf{Left}: Four skeletons dancing in the same manner were manually segmented and tracked during a single one second interval. 
\textbf{Center}: Affinity matrix before binarization. The slight differences in the entries are due to the intrinsic variability of the action itself. \textbf{Right}: Binarized affinity matrix after applying the threshold $\tau = 0.9/3= 0.3$. The values in the entries show slight variation but all larger than $0.3$, so they are all binarized to 1, and thus the four skeletons are grouped together.}
\label{fig:skeleton_dancing_example}

\end{figure}

We now test the classification performance of the framework described in Section~\ref{sec:space} for grouping actions per time interval. The cartoon-skeletons dancing in the Skeleton video, as shown in  Fig.~\ref{fig:skeleton_dancing_example}, were segmented and tracked manually to illustrate that individual actions have intrinsic variability, even when the actions performed are the same, justifying the relaxation coefficient $r$. Notice that this effect is not a by-product of the tracking/segmentation procedure, since it is done manually in this example.

\begin{figure}[ht]
   \centering   
   \includegraphics[width=0.15\textwidth]{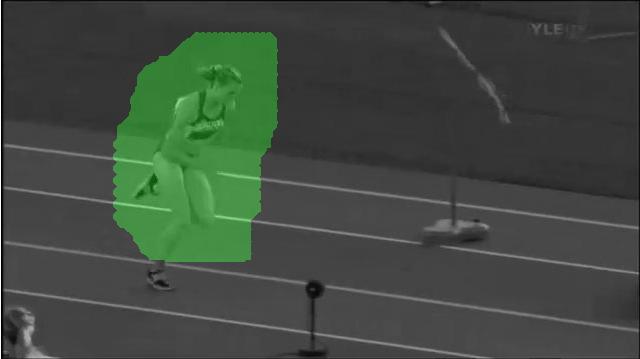}
   \includegraphics[width=0.15\textwidth]{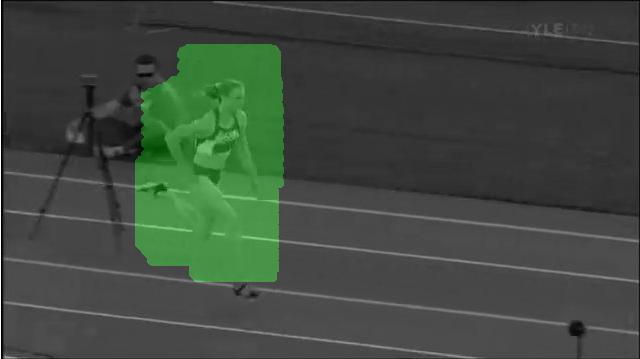} 
   \includegraphics[width=0.15\textwidth]{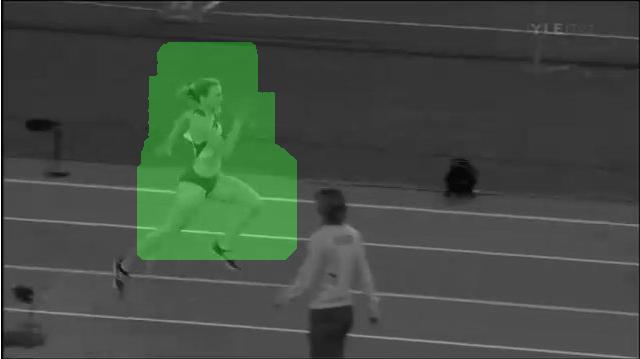}
   \includegraphics[width=0.15\textwidth]{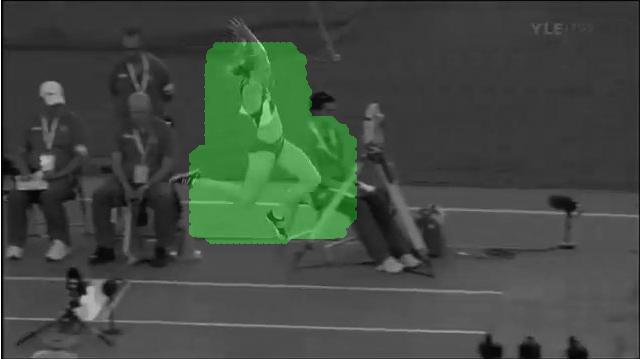}
   \includegraphics[width=0.15\textwidth]{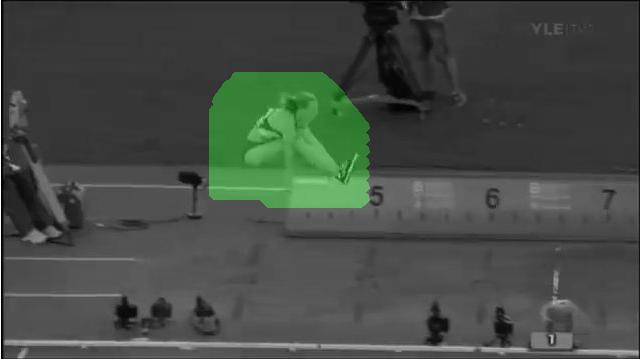} 
   \includegraphics[width=0.15\textwidth]{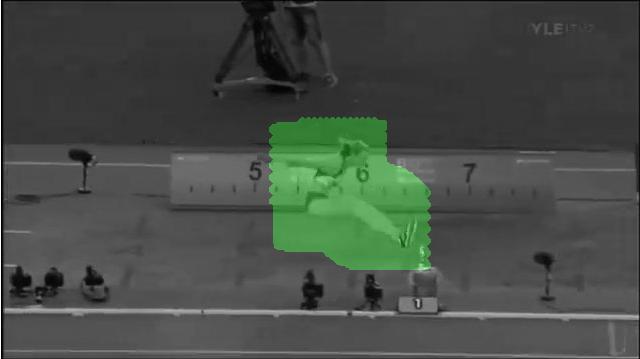} \\
   \includegraphics[width=0.15\textwidth]{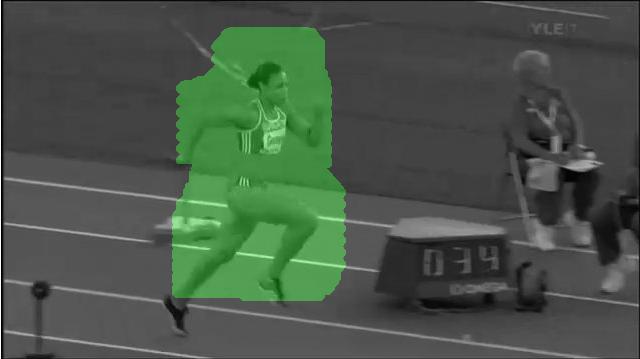} 
   \includegraphics[width=0.15\textwidth]{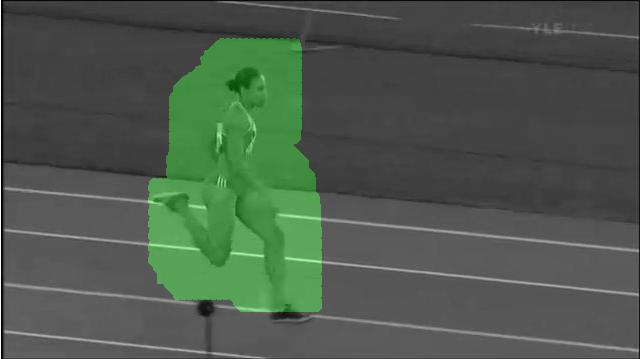} 
   \includegraphics[width=0.15\textwidth]{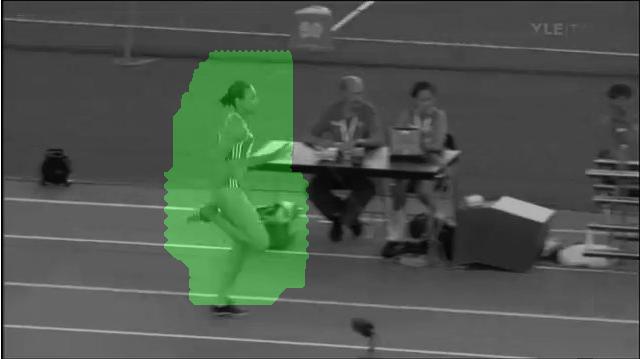} 
   \includegraphics[width=0.15\textwidth]{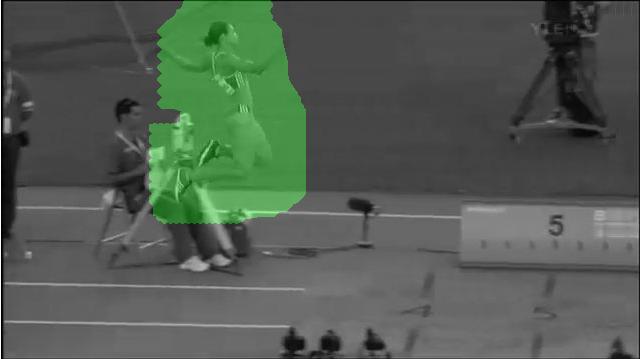} 
   \includegraphics[width=0.15\textwidth]{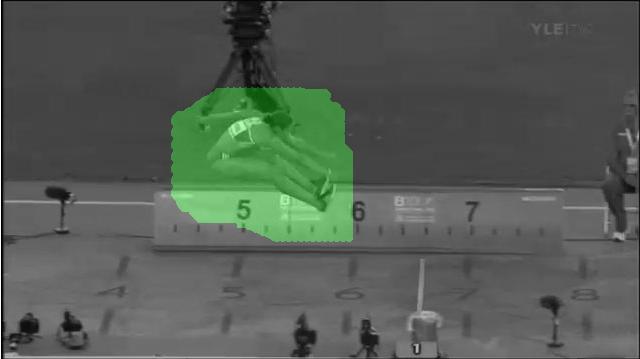} 
   \includegraphics[width=0.15\textwidth]{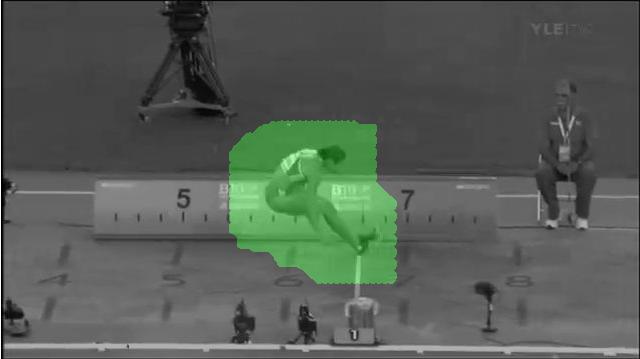} \\
   \includegraphics[width=0.15\textwidth]{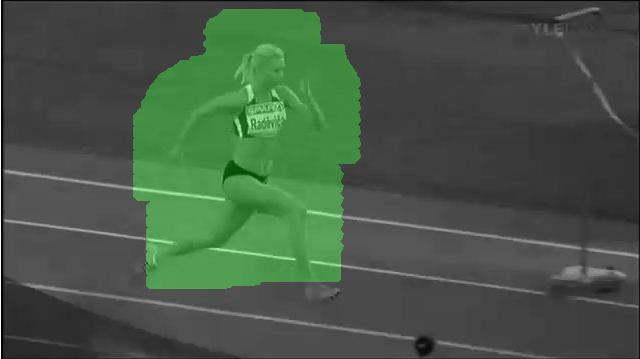}
   \includegraphics[width=0.15\textwidth]{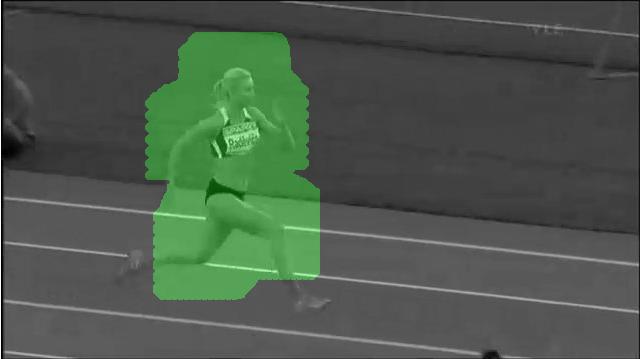} 
   \includegraphics[width=0.15\textwidth]{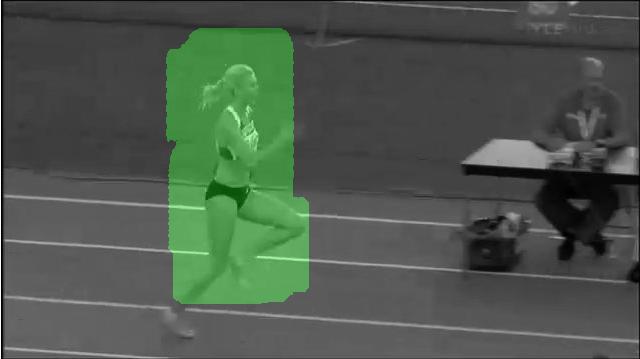} 
   \includegraphics[width=0.15\textwidth]{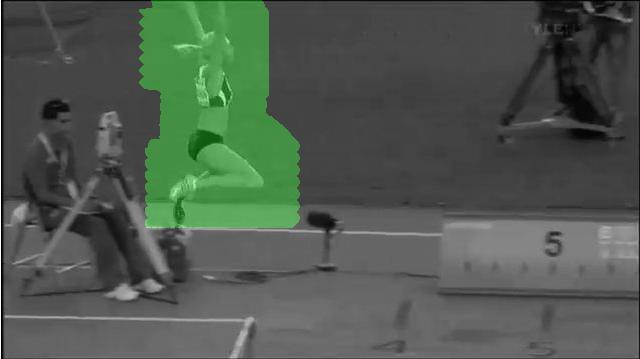} 
   \includegraphics[width=0.15\textwidth]{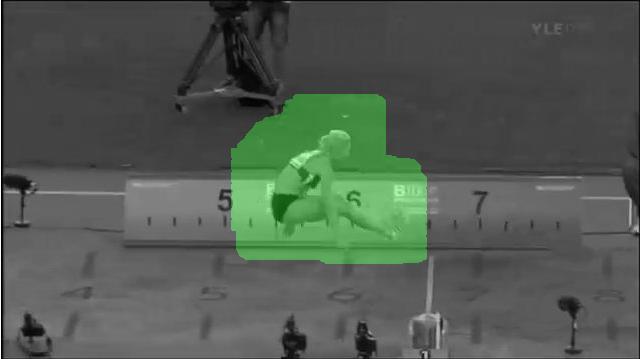} 
   \includegraphics[width=0.15\textwidth]{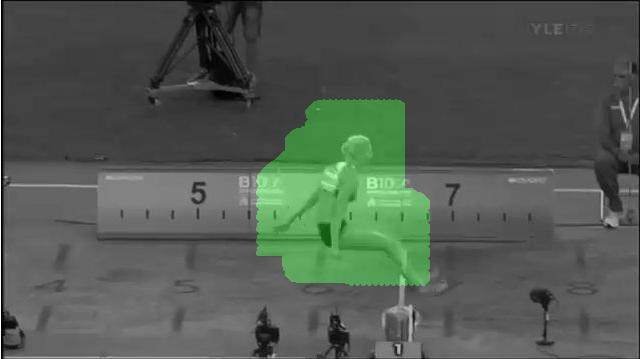}   
  \caption{Sample frames from the Long jump video.  On each row, an athlete is running and then long-jumping.  Colored segmentation masks are displayed.
  (This is a color figure.)}
  \label{fig:imgs_running_long_jumping}
\end{figure}

\setlength\fboxsep{0pt}
\setlength\fboxrule{1.0pt}
We then analyzed the Long jump video, shown in Fig.~\ref{fig:imgs_running_long_jumping}, where three persons are performing two actions: running and long-jumping. 
We tested every possible 
configuration of three video individuals (with corresponding time segments cropped from the original video), showing that we are doing action classification and not person classification. These results are summarized in Table~\ref{table:running_long_jumping}, where we obtained one single grouping error (see the third configuration).

\begin{table}[ht]

\caption{Three people running (R) and then long-jumping (LJ). 
The grouping decision is represented with letters A and B, with only one grouping error on the third configuration (cells are colored to facilitate the comparison between the results and the ground truth, matching colors means correct result). See  Fig.~\ref{fig:imgs_running_long_jumping} for sample frames from this video.}
\label{table:running_long_jumping}

\centering
\begin{tabularx}{\textwidth}{>{\centering}p{1.5cm}<{\centering}*{9}{C}}
\toprule
           & \textbf{Persons} & \multicolumn{8}{c}{\textbf{Action grouping}} \\
\midrule
\multirow{3}{2cm}{Ground\\Truth}
& I & \actionB{R} & \actionA{LJ} & \actionA{LJ} & \actionB{R} & \actionB{R} & \actionB{R} & \actionA{LJ} & \actionA{LJ} \\
& II & \actionB{R} & \actionA{LJ} & \actionB{R} & \actionA{LJ} & \actionB{R} & \actionA{LJ} & \actionB{R} & \actionA{LJ}\\
& III & \actionB{R} & \actionA{LJ} & \actionB{R} & \actionB{R} & \actionA{LJ} & \actionA{LJ} & \actionA{LJ} & \actionB{R}\\
\addlinespace
\multirow{3}{2cm}{Result}
& I & \actionB{A} & \actionA{A} & \actionB{A} & \actionB{A} & \actionB{A} & \actionB{A} & \actionA{A} & \actionA{A}  \\ 
& II & \actionB{A} & \actionA{A} & \actionB{A} & \actionA{B} & \actionB{A} & \actionA{B} & \actionB{B} & \actionA{A} \\ 
& III & \actionB{A} & \actionA{A} & \actionB{A} & \actionB{A} & \actionA{B} & \actionA{B} & \actionA{A} & \actionB{B} \\ 
\bottomrule 
\end{tabularx} 

\end{table}

\begin{figure}[ht]
  \centering 
  \includegraphics[width=0.24\textwidth, height=0.12\textheight]{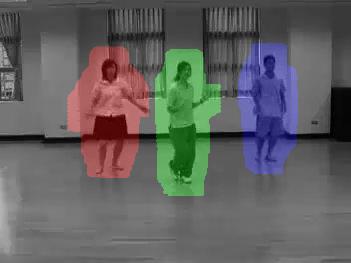} 
  \includegraphics[width=0.24\textwidth, height=0.12\textheight]{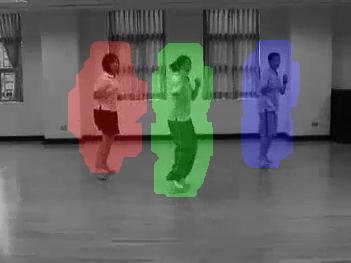}   
  \includegraphics[width=0.24\textwidth, height=0.12\textheight]{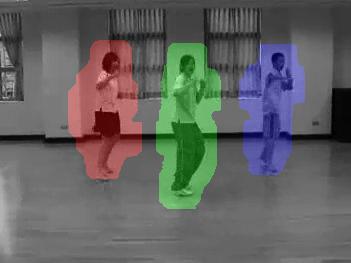}
  \includegraphics[width=0.24\textwidth, height=0.12\textheight]{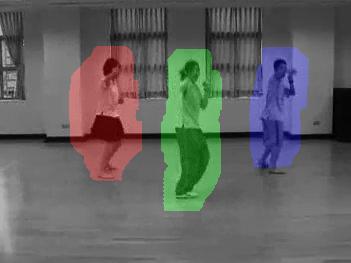} \\
  \includegraphics[width=0.24\textwidth, height=0.12\textheight]{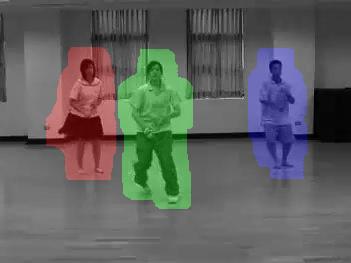} 
  \includegraphics[width=0.24\textwidth, height=0.12\textheight]{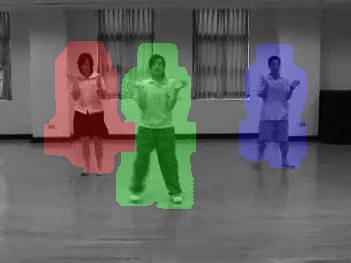}
  \includegraphics[width=0.24\textwidth, height=0.12\textheight]{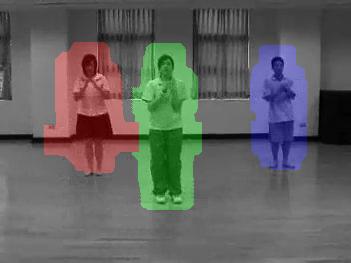}
  \includegraphics[width=0.24\textwidth, height=0.12\textheight]{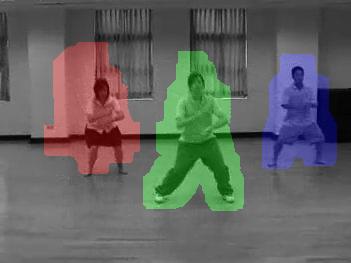}      
  \caption{Sample frames from the Gym video.  On the first row, the three individuals are punching; on the second row, they are dancing. 
  The segmentation masks for the different individuals appear in different colors. (This is a color figure.)}
  \label{fig:imgs_punching_dancing}
\end{figure}

The test for the Gym video, shown in Fig.~\ref{fig:imgs_punching_dancing}, consists of 
three persons performing two actions: punching and dancing. 
The results are shown in Table~\ref{table:punching_dancing}.  In this test, we again obtained only one grouping error.

\begin{figure}[!h]
  \centering 
  \includegraphics[width=0.24\textwidth, height=0.12\textheight]{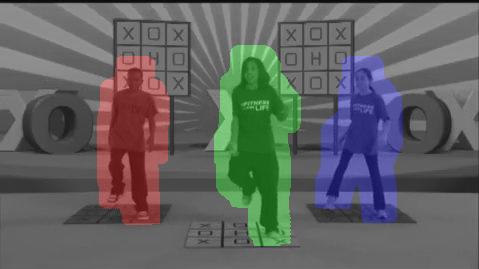} 
  \includegraphics[width=0.24\textwidth, height=0.12\textheight]{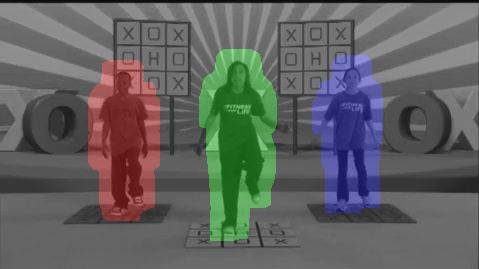}   
  \includegraphics[width=0.24\textwidth, height=0.12\textheight]{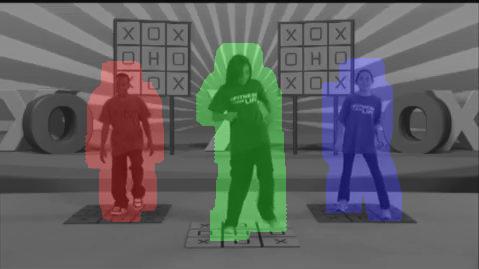}
  \includegraphics[width=0.24\textwidth, height=0.12\textheight]{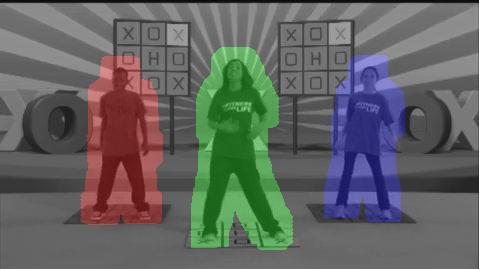} \\
  \includegraphics[width=0.24\textwidth, height=0.12\textheight]{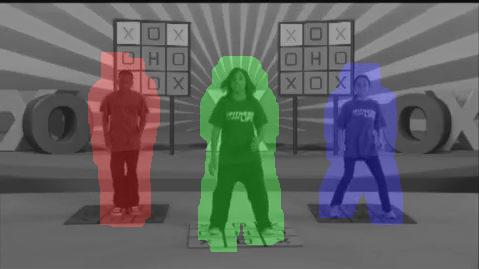} 
  \includegraphics[width=0.24\textwidth, height=0.12\textheight]{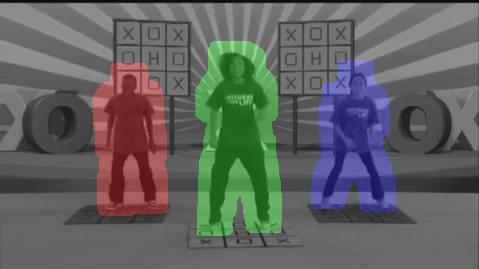}
  \includegraphics[width=0.24\textwidth, height=0.12\textheight]{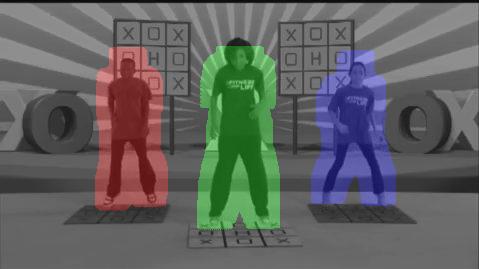}
  \includegraphics[width=0.24\textwidth, height=0.12\textheight]{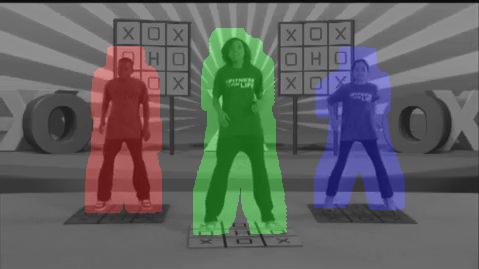}    
  \caption{Sample frames from the Kids video. Fist row: dancing. Second row: jumping.
  The segmentation masks for the different individuals appear in different colors. (This is a color figure.)}
  \label{fig:imgs_kids}
\end{figure}

\begin{table}[ht]

\caption{Analysis of the Gym video. Three persons are punching (P) or dancing (D) (cells are colored to facilitate the comparison).
The grouping decision is shown with values A or B, with only one grouping error.
See Fig.~\ref{fig:imgs_punching_dancing}  for some typical frames.}
\label{table:punching_dancing}

\centering
\begin{tabularx}{\textwidth}{>{\centering}p{1.5cm}<{\centering}*{9}{C}}
\toprule
 & \textbf{Persons} & \multicolumn{8}{c}{\textbf{Action Grouping}} \\
\midrule
\multirow{3}{2cm}{Ground\\Truth}
& I & \actionA{P} & \actionB{D} & \actionA{P} & \actionB{D} & \actionB{D} & \actionB{D} & \actionA{P} & \actionA{P}\\
& II & \actionA{P} & \actionB{D} & \actionB{D} & \actionA{P} & \actionB{D} & \actionA{P} & \actionB{D} & \actionA{P}\\
& III & \actionA{P} & \actionB{D} & \actionB{D} & \actionB{D} & \actionA{P} & \actionA{P} & \actionA{P} & \actionB{D}\\
\addlinespace
\multirow{3}{2cm}{Result}
& I & \actionA{A} & \actionB{A} &\actionA{A} & \actionB{A} & \actionB{A} & \actionA{A} & \actionA{A} & \actionA{A} \\ 
& II & \actionA{A} & \actionB{A} & \actionB{B} & \actionA{B} & \actionB{A} & \actionA{A} & \actionB{B} & \actionA{A} \\ 
& III & \actionA{A} & \actionB{A} & \actionB{B} & \actionB{A} & \actionA{B} & \actionA{A} & \actionA{A} & \actionB{B} \\ 
\bottomrule
\end{tabularx} 


\end{table}

To further validate that we are doing action classification (not just classifying the person itself), and that we can treat more imbalanced cases (most individuals performing an action), we also conducted a `cloning' test, using the Gym video (Fig.~\ref{fig:imgs_punching_dancing}).
In this scenario, we added a fourth person by artificially replicating one of the original three, but performing (in a different video segment) an action different than the one of the original time interval. The results are shown in Table~\ref{table:four_persons_punching_dancing}. 
A correct grouping result was attained, confirming that the proposed method only perceives actions, and is robust to differences in human appearance. 

\begin{table}[ht]

\caption{Analysis of the Gym video. Three persons are punching (P) or dancing (D), and a `clone' is added.
The second column denotes the person's index.
The fourth person, that is the clone, is the same as one of the first three, but is doing something different. For example, 
I-D means that the person I was cloned. 
The (perfect) grouping decision is shown inside the table (cells are colored to facilitate the comparison).
See Fig.~\ref{fig:imgs_punching_dancing}  for some typical frames.}
\label{table:four_persons_punching_dancing}

\centering
\begin{tabularx}{\textwidth}{>{\centering}p{1.5cm}p{1.5cm}<{\centering}*{7}{C}}
\toprule
                          &
\textbf{Persons} &
\multicolumn{7}{c}{\textbf{Action Grouping}} \\
\midrule
\multirow{4}{1cm}{Ground\\Truth} &
I & \actionA{P} & \actionA{P} & \actionA{P} & \actionA{P} & \actionB{D}  & \actionB{D}  & \actionB{D}\\
& II & \actionA{P} & \actionA{P} & \actionA{P} & \actionA{P} & \actionB{D}  & \actionB{D}  & \actionB{D}\\
& III & \actionB{D} & \actionA{P} & \actionA{P} & \actionA{P} & \actionB{D}  & \actionB{D}  & \actionB{D}\\
& `Clone' & \actionB{II-D} & \actionB{I-D} & \actionB{II-D} & \actionB{III-D} & \actionA{I-P}  & \actionA{II-P}  & \actionA{III-P}\\
\addlinespace
\multirow{4}{*}{Result} &
I & \actionA{A} & \actionA{A} & \actionA{A} & \actionA{A} & \actionB{A}  & \actionB{A}  & \actionB{A}\\
& II & \actionA{A} & \actionA{A} & \actionA{A} & \actionA{A} & \actionB{A}  & \actionB{A}  & \actionB{A}\\
& III & \actionB{B} & \actionA{A} & \actionA{A} & \actionA{A} & \actionB{A}  & \actionB{A}  & \actionB{A}\\
& `Clone' & \actionB{B} & \actionB{B} & \actionB{B} & \actionB{B} & \actionA{B}  & \actionA{B}  & \actionA{B}\\
\bottomrule

\end{tabularx}

\end{table}

\begin{figure}[ht]
\centering
\begin{tabular}{@{\hspace{0pt}}m{.27\textwidth}*{2}{@{\hspace{4pt}}m{.33\textwidth}}@{\hspace{0pt}}}
\begin{minipage}{.27\textwidth}
\centering
\includegraphics[width=\textwidth]{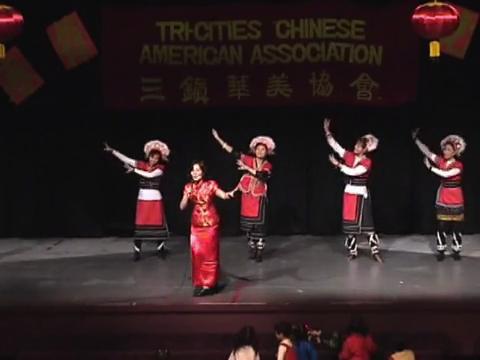}\\
\includegraphics[width=\textwidth]{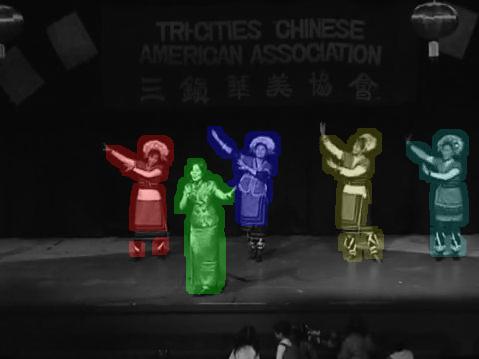} 
\end{minipage} &
\includegraphics[width=.33\textwidth]{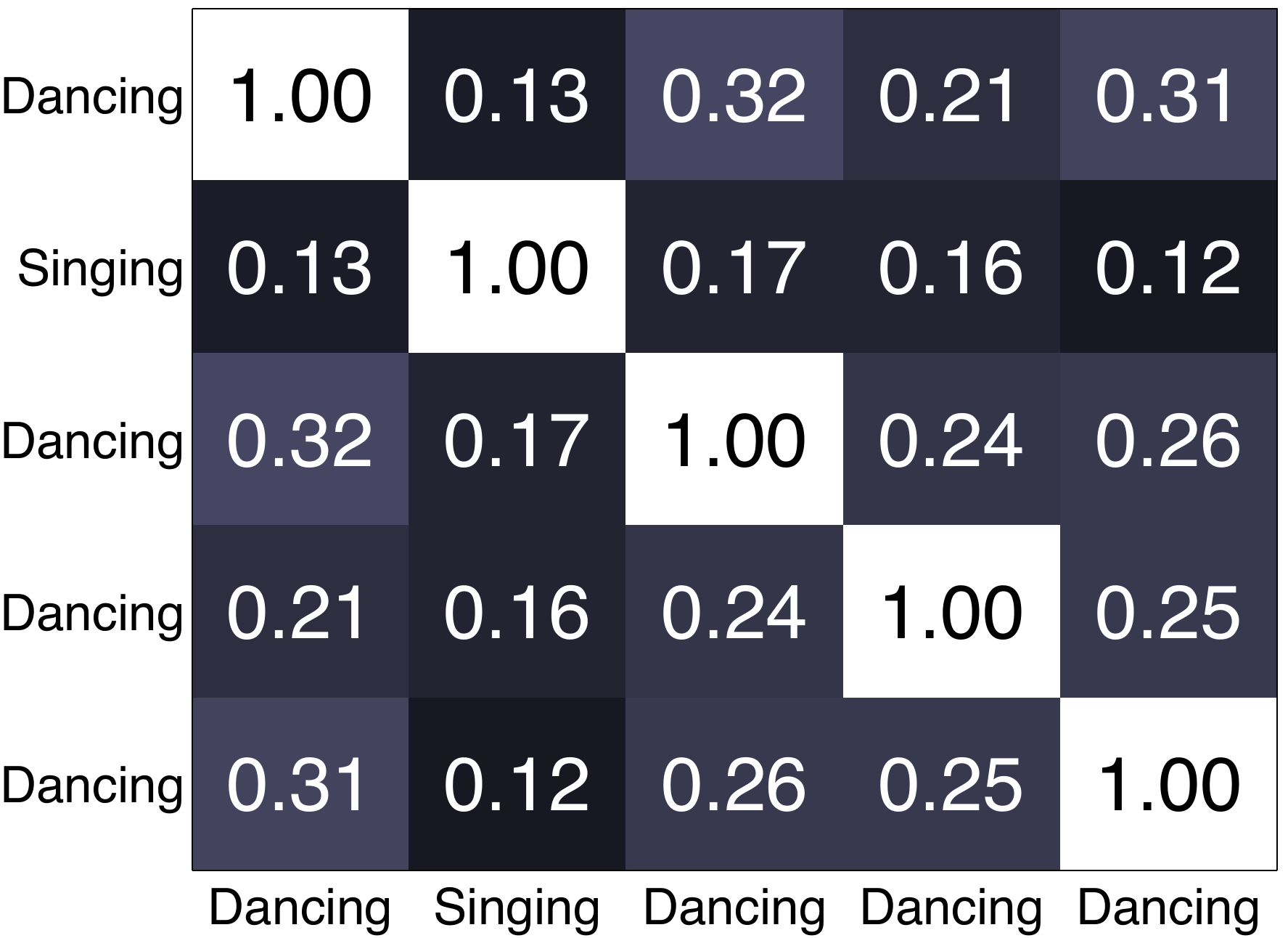} & 
\includegraphics[width=.33\textwidth]{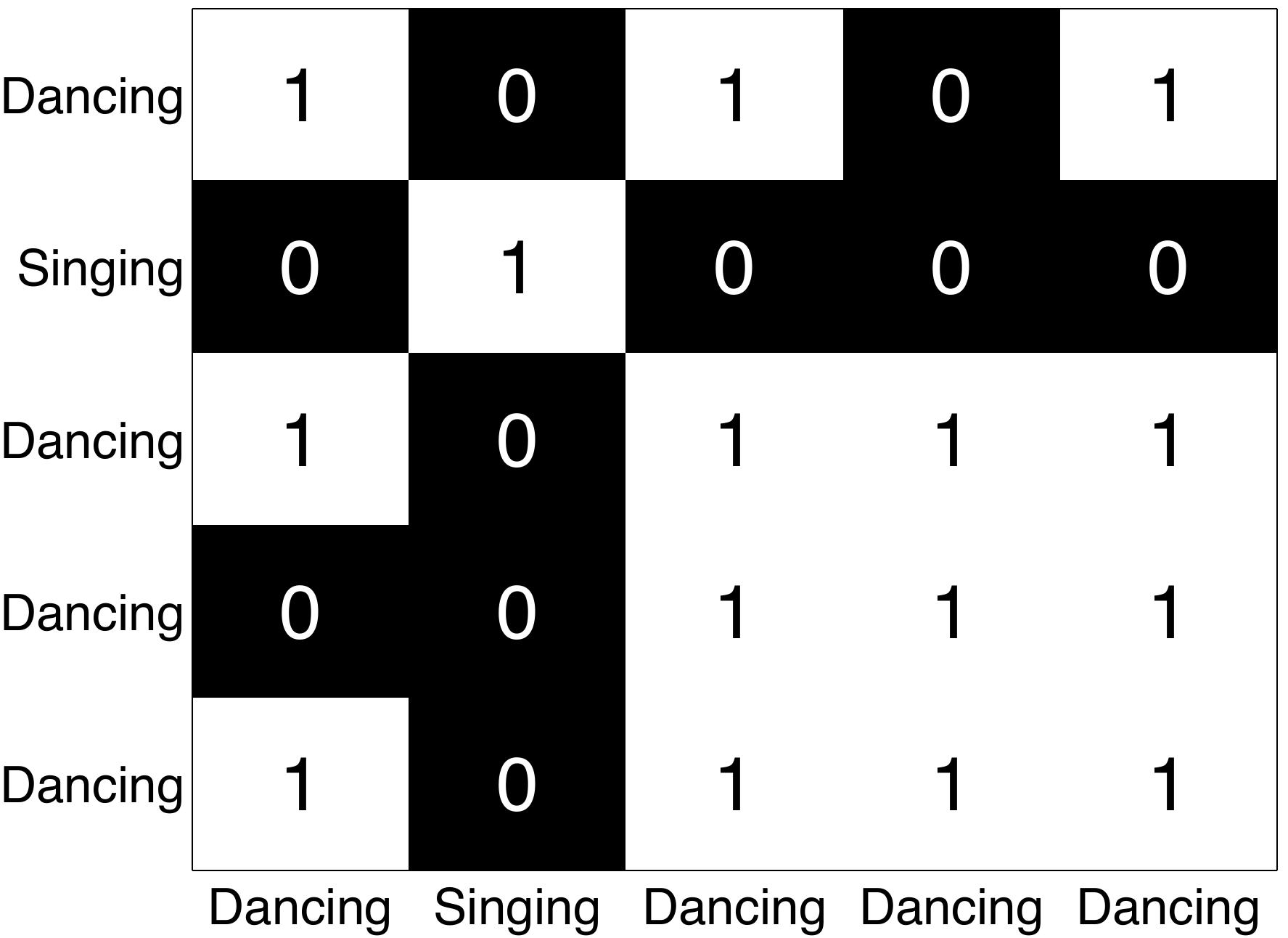} \\
\end{tabular}
\caption{Sample frames from the Singing-dancing video.  \textbf{Left}: Five persons, a singer and four dancers, were tracked/segmented during a one second interval (the masks are displayed in colors on the bottom left, the singer appearing in green). \textbf{Center}: Affinity matrix. \textbf{Right}: Binarized affinity matrix after applying the threshold $\tau = 0.9/4 = 0.225$. Note that the values in the entries of the second row and the second column (except
the diagonal entries) are small, hence binarized to zero. This implies that the second
person is doing a different action than the group. The binarization on entries (1,4)
and (4,1) fails to be 1, not affecting the grouping, since the persons are grouped as connected components. 
Two groups are correctly detected, the four dancers and the singer.
(This is a color figure.)}
\label{fig:taiwanese_dancing_example}
\end{figure}

\begin{figure}[ht]
\centering
\begin{tabular}{@{\hspace{0pt}}m{.27\textwidth}*{2}{@{\hspace{4pt}}m{.33\textwidth}}@{\hspace{0pt}}}
\begin{minipage}{.27\textwidth}
\includegraphics[width=\textwidth]{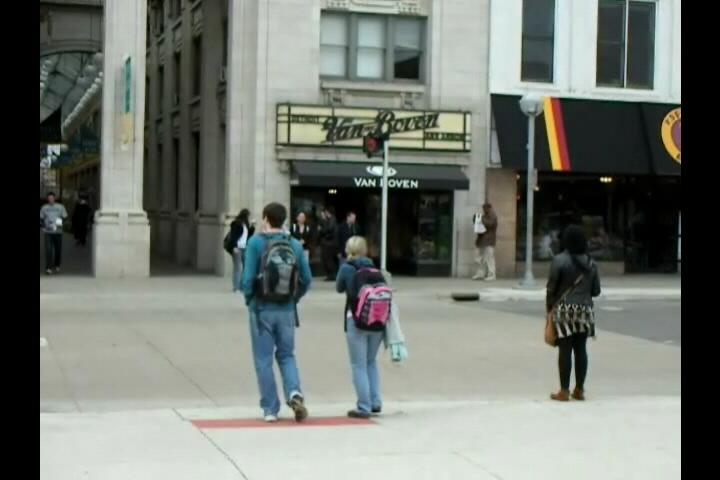}\\
\vspace{-8pt}
\includegraphics[width=\textwidth]{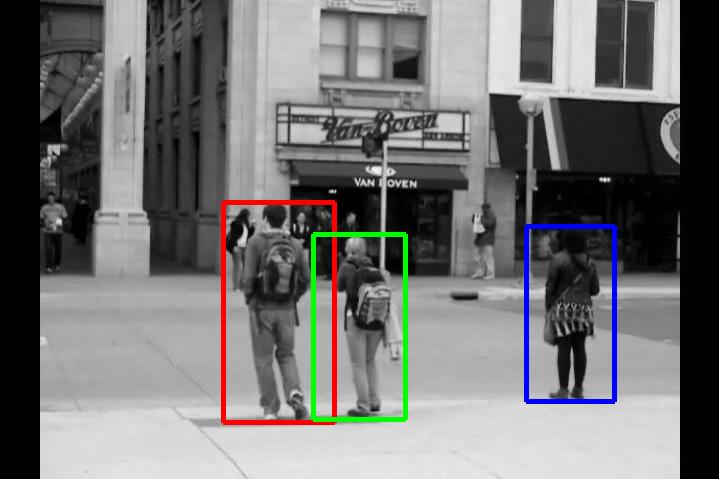} 
\end{minipage} &
\includegraphics[width=.33\textwidth]{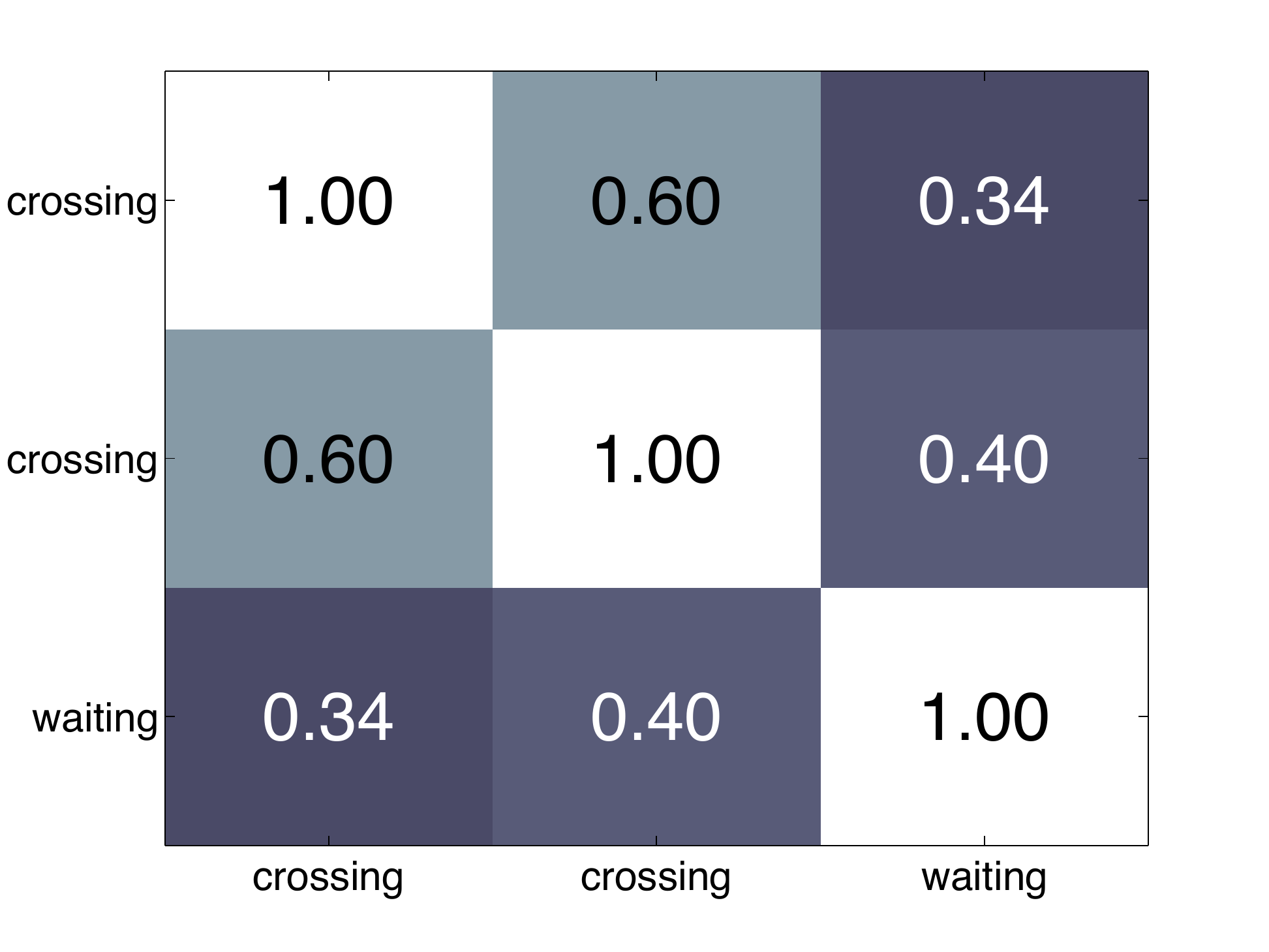} & 
\includegraphics[width=.33\textwidth]{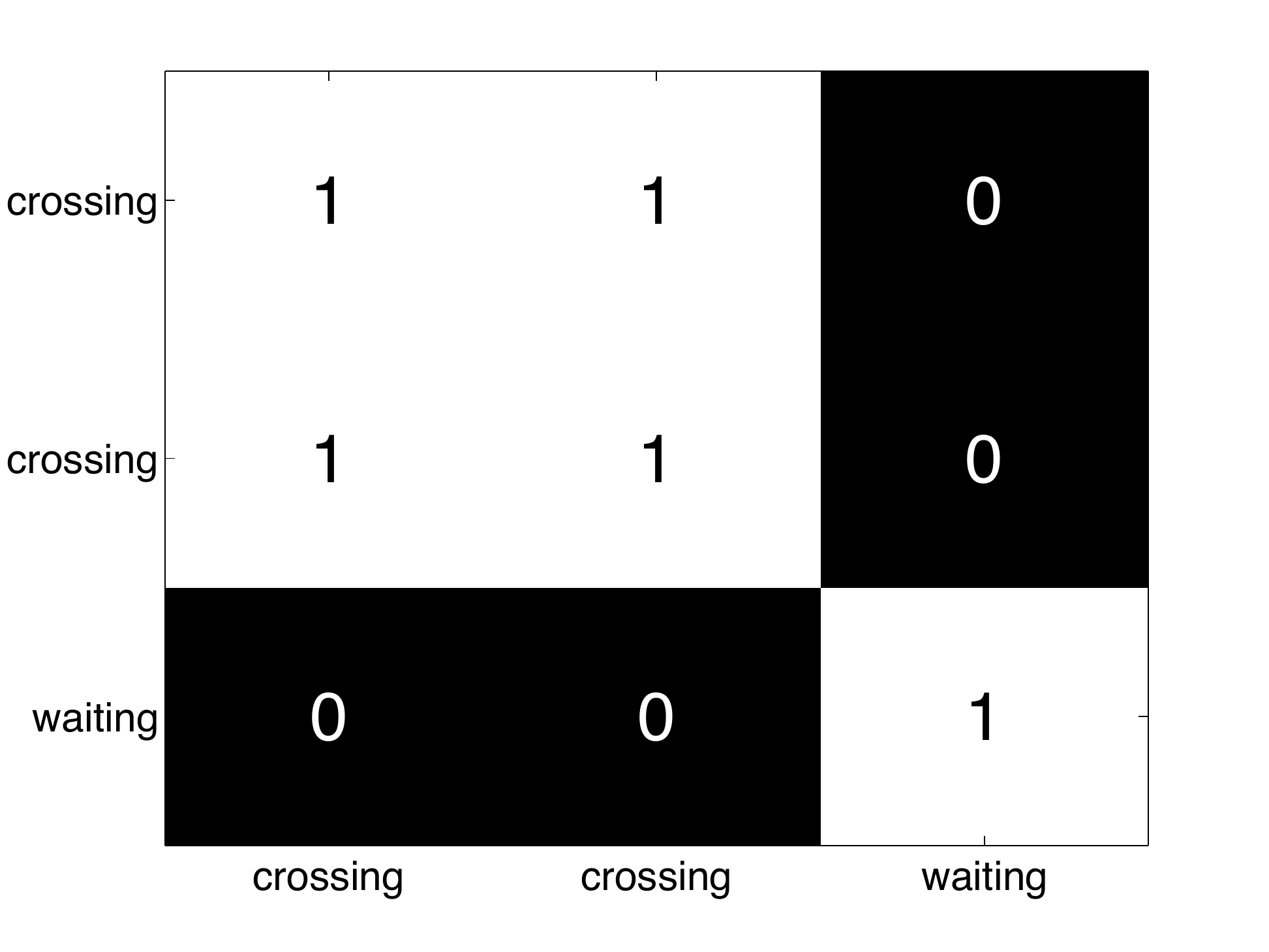} \\
\end{tabular}
\caption{\textbf{Left}: A sample frame of the Crossing video. Two pedestrians at left are crossing the street while the one at right is waiting. The provided bounding boxes were used instead of running the 
tracking/segmentation algorithm. \textbf{Center}: Affinity matrix. \textbf{Right}: Binarized affinity matrix after applying the threshold $\tau = 0.9/2= 0.45$. Two groups are correctly detected. 
(This is a color figure.)}
\label{fig:crossing_waiting_example}
\end{figure}

\begin{figure}[ht]
\centering
\begin{tabular}{@{\hspace{0pt}}m{.27\textwidth}*{2}{@{\hspace{4pt}}m{.33\textwidth}}@{\hspace{0pt}}}
\begin{minipage}{.27\textwidth}
\includegraphics[width=\textwidth]{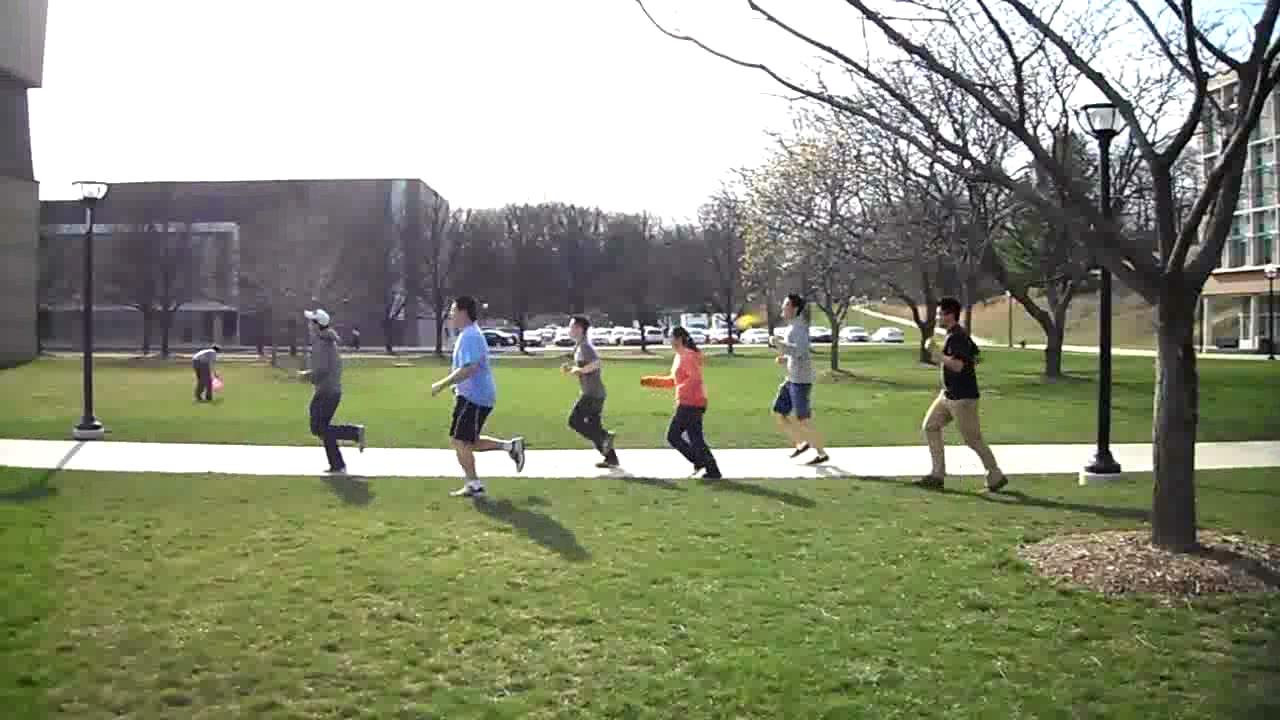}\\
\vspace{-8pt}
\includegraphics[width=\textwidth]{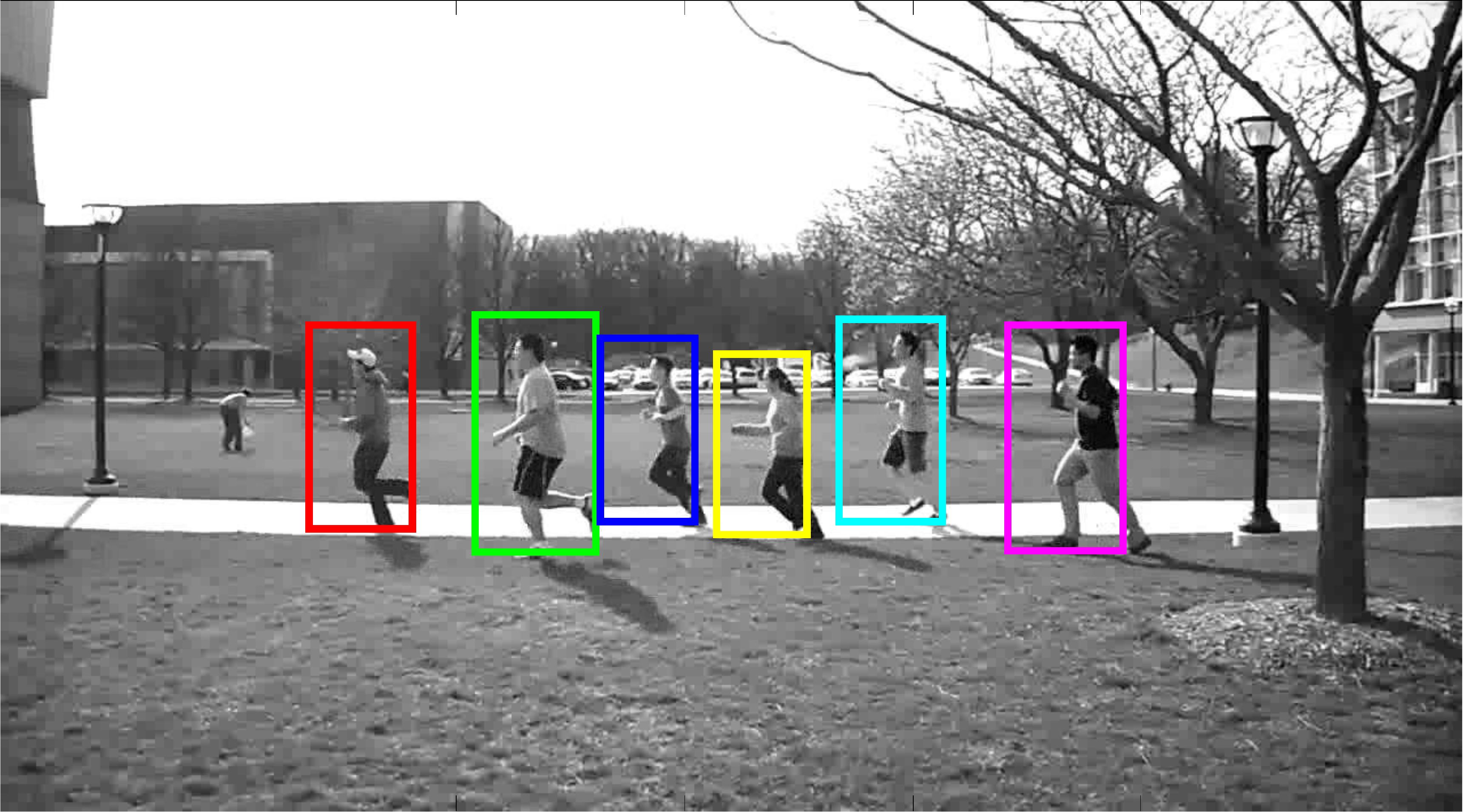} 
\end{minipage} &
\includegraphics[width=.33\textwidth]{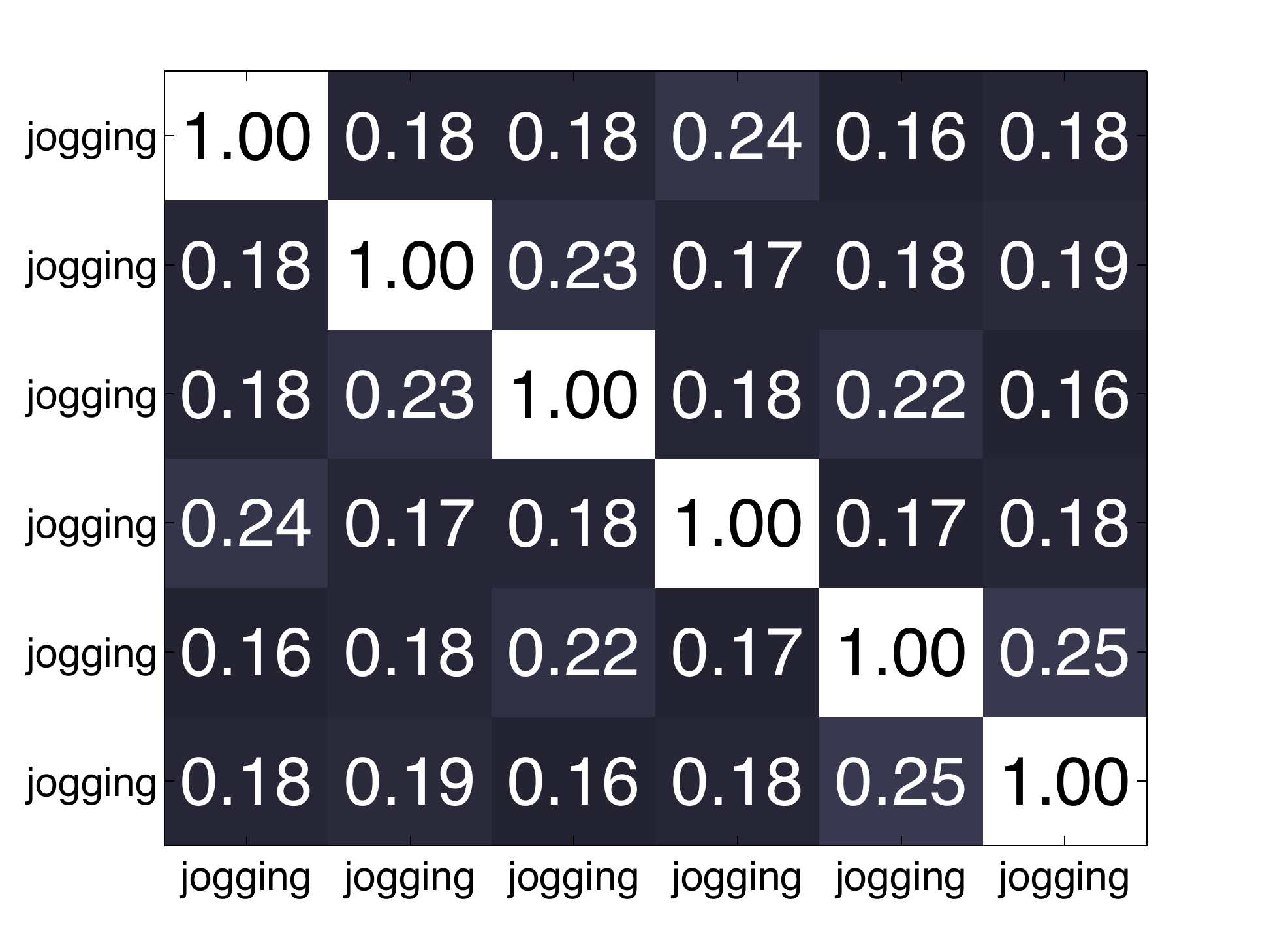} & 
\includegraphics[width=.33\textwidth]{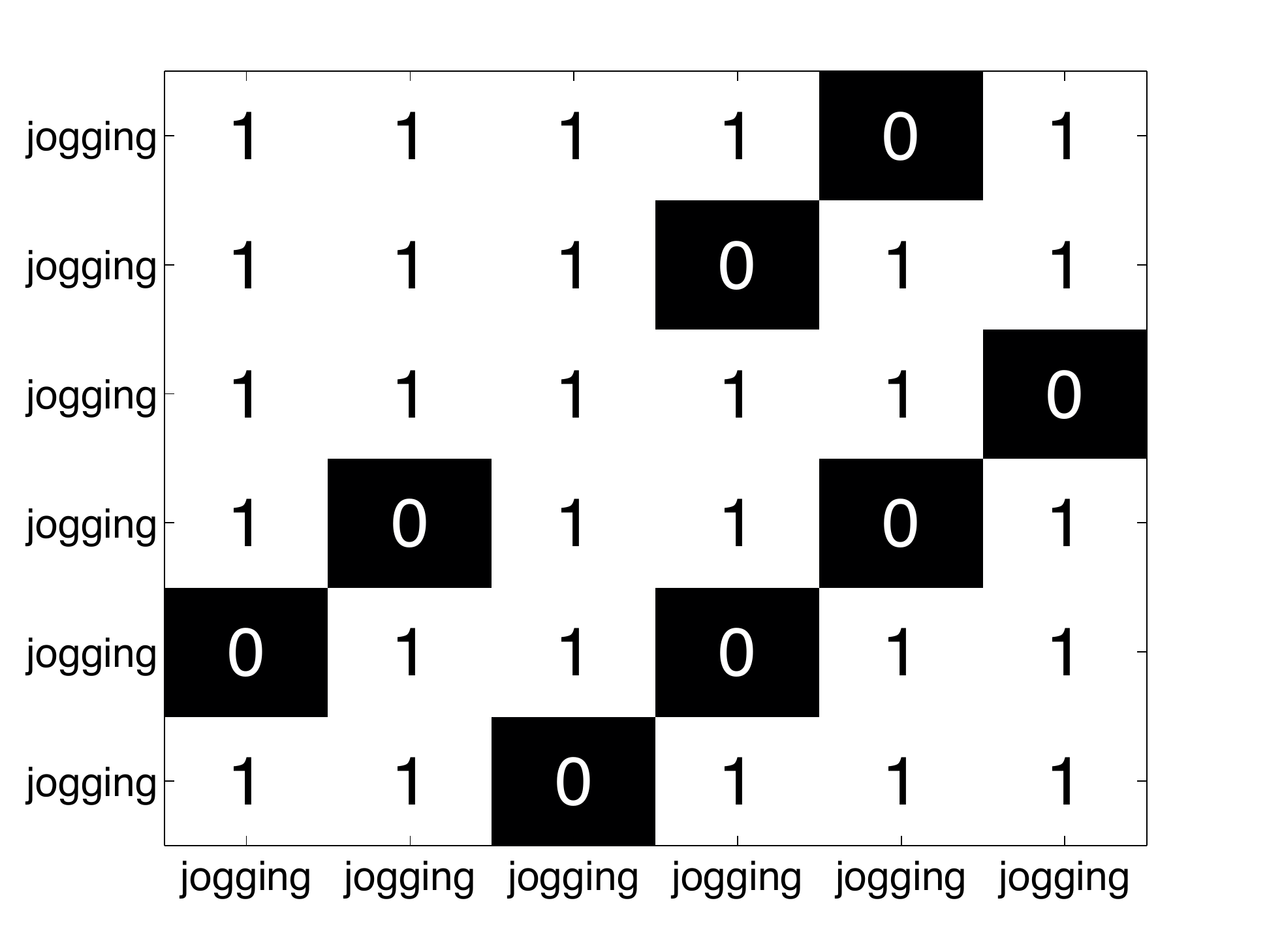} \\
\end{tabular}
\caption{\textbf{Left}: A sample frame of the Jogging video. The provided bounding boxes were used instead of running the 
tracking/segmentation algorithm. \textbf{Center}: Affinity matrix. \textbf{Right}: Binarized affinity matrix after applying the threshold $\tau = 0.9/5 = 0.18$. Note that some entries are a little smaller than the threshold $0.18$, thus binarized to be $0$. 
But the grouping result is still correct since persons are grouped as connected components. One single group is correctly detected. 
(This is a color figure.)}
\label{fig:jogging_example}
\end{figure}

\begin{figure}[ht]
\centering
\begin{tabular}{@{\hspace{0pt}}m{.27\textwidth}*{2}{@{\hspace{4pt}}m{.33\textwidth}}@{\hspace{0pt}}}
\begin{minipage}{.27\textwidth}
\includegraphics[width=\textwidth]{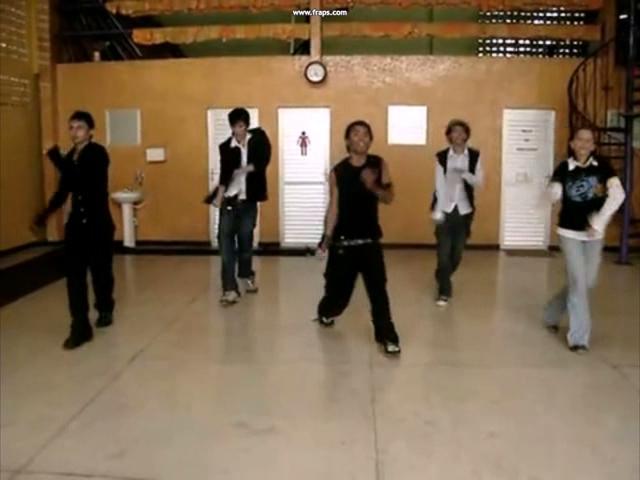}\\
\vspace{-8pt}
\includegraphics[width=\textwidth]{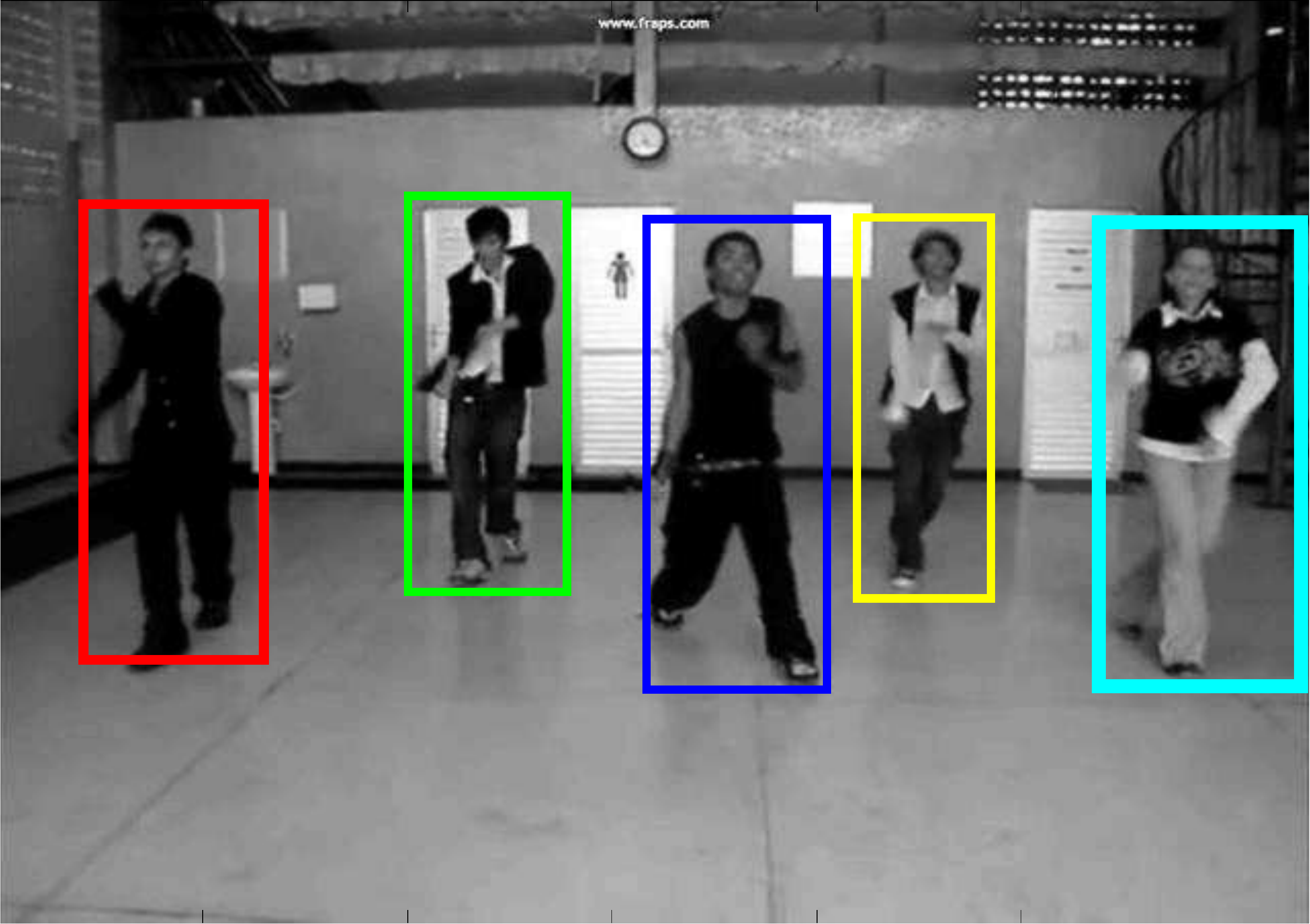} 
\end{minipage} &
\includegraphics[width=.33\textwidth]{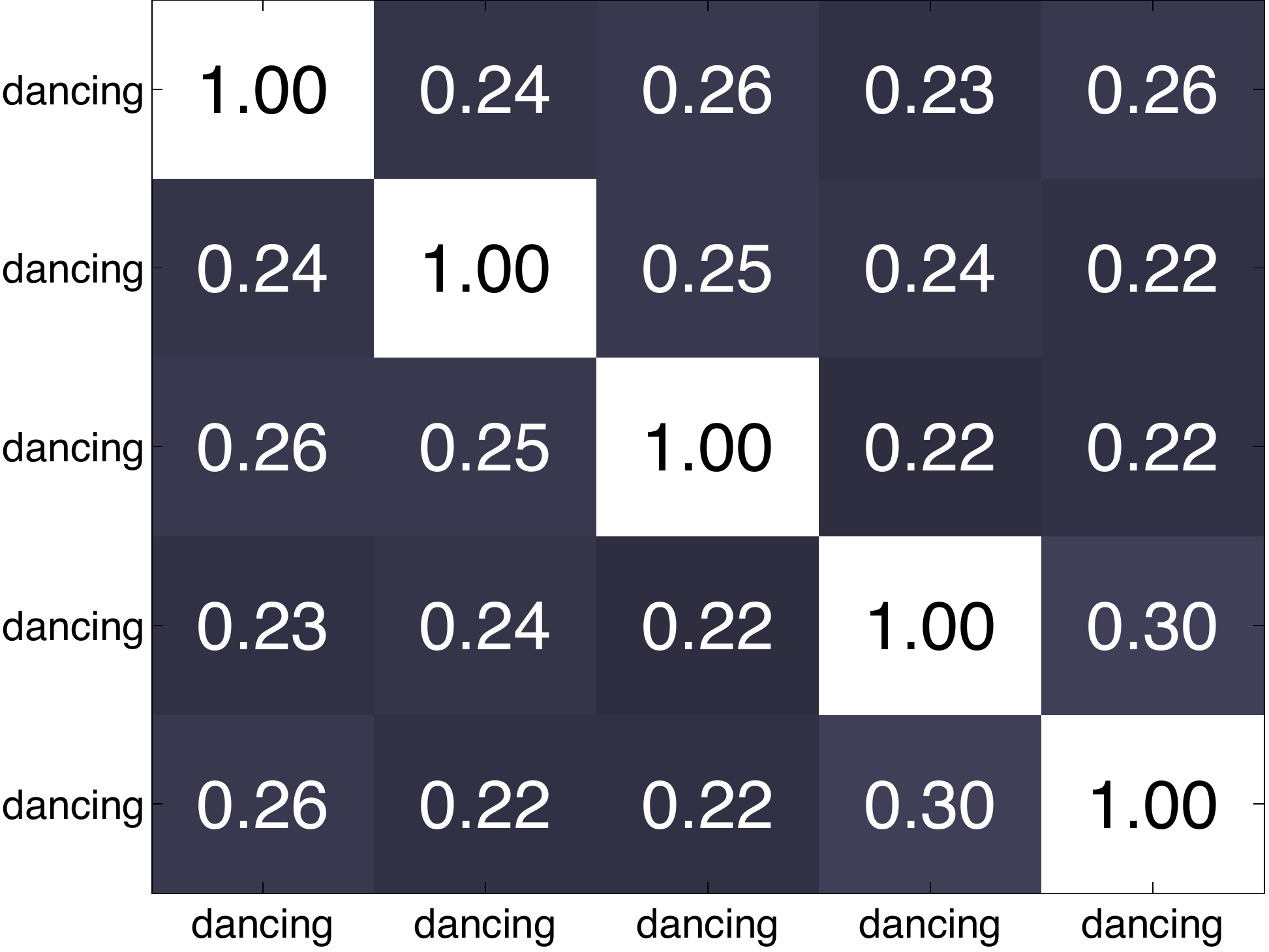} & 
\includegraphics[width=.33\textwidth]{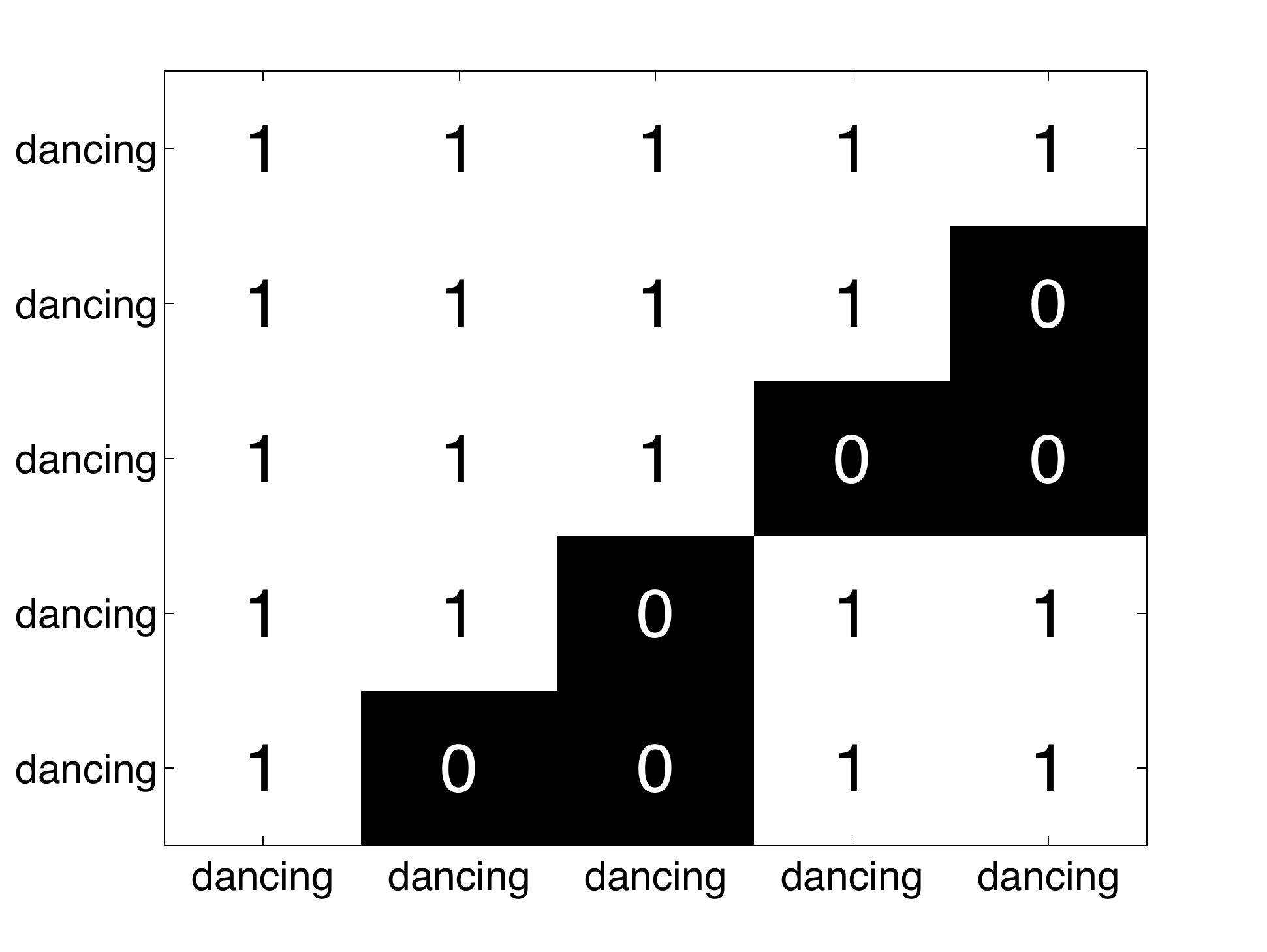} \\
\end{tabular}
\caption{\textbf{Left}: A sample frame of the Dancing video. The provided bounding boxes were used instead of running the 
tracking/segmentation algorithm. \textbf{Center}: Affinity matrix. \textbf{Right}: Binarized affinity matrix after applying the threshold $\tau = 0.9/4 = 0.225$. Note that some entries are a little smaller than the threshold $0.225$, thus binarized to be $0$. 
But the grouping result is still correct since persons are grouped as connected components. One single group is correctly detected. 
(This is a color figure.)}
\label{fig:dancing_example}
\end{figure}

On a more imbalanced scenario, we analyzed the Singing-dancing video, where five individuals are dancing while the remaining one is singing, see Fig.~\ref{fig:taiwanese_dancing_example}. 
From the affinity matrix, 
we observe that the row and column corresponding to the second (singing) person have smaller values, binarized to zero after applying the thresholding operation. 
The threshold in this case is $\tau = 0.9 / 4 = 0.225$, slightly larger than the values $0.21$ in entries $(1, 4)$ and $(4, 1)$, which should be 
binarized to 1. Since the persons are grouped as connected components, we still obtain the correct groups. 
More imbalanced tests for the Jogging, Dancing and Crossing videos (shown in figures ~\ref{fig:crossing_waiting_example}, \ref{fig:jogging_example} and \ref{fig:dancing_example}) also give correct grouping results. 

\begin{figure}[ht]
\centering

\subfloat[A single group is correctly identified.]{
    \begin{tabular}{@{\hspace{0pt}}m{.27\textwidth}*{2}{@{\hspace{4pt}}m{.33\textwidth}}@{\hspace{0pt}}}
    \begin{minipage}{.27\textwidth}
    \centering
    \includegraphics[width=\textwidth]{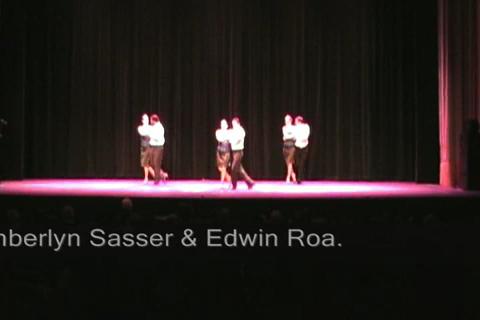}\\
    \includegraphics[width=\textwidth]{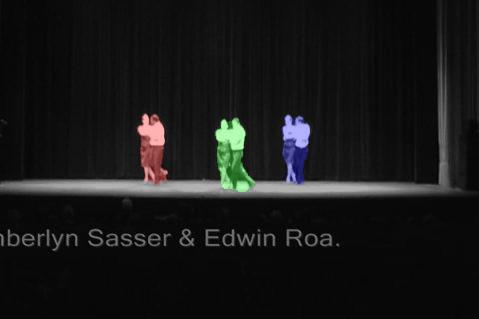} 
    \end{minipage} &
    \includegraphics[width=.33\textwidth]{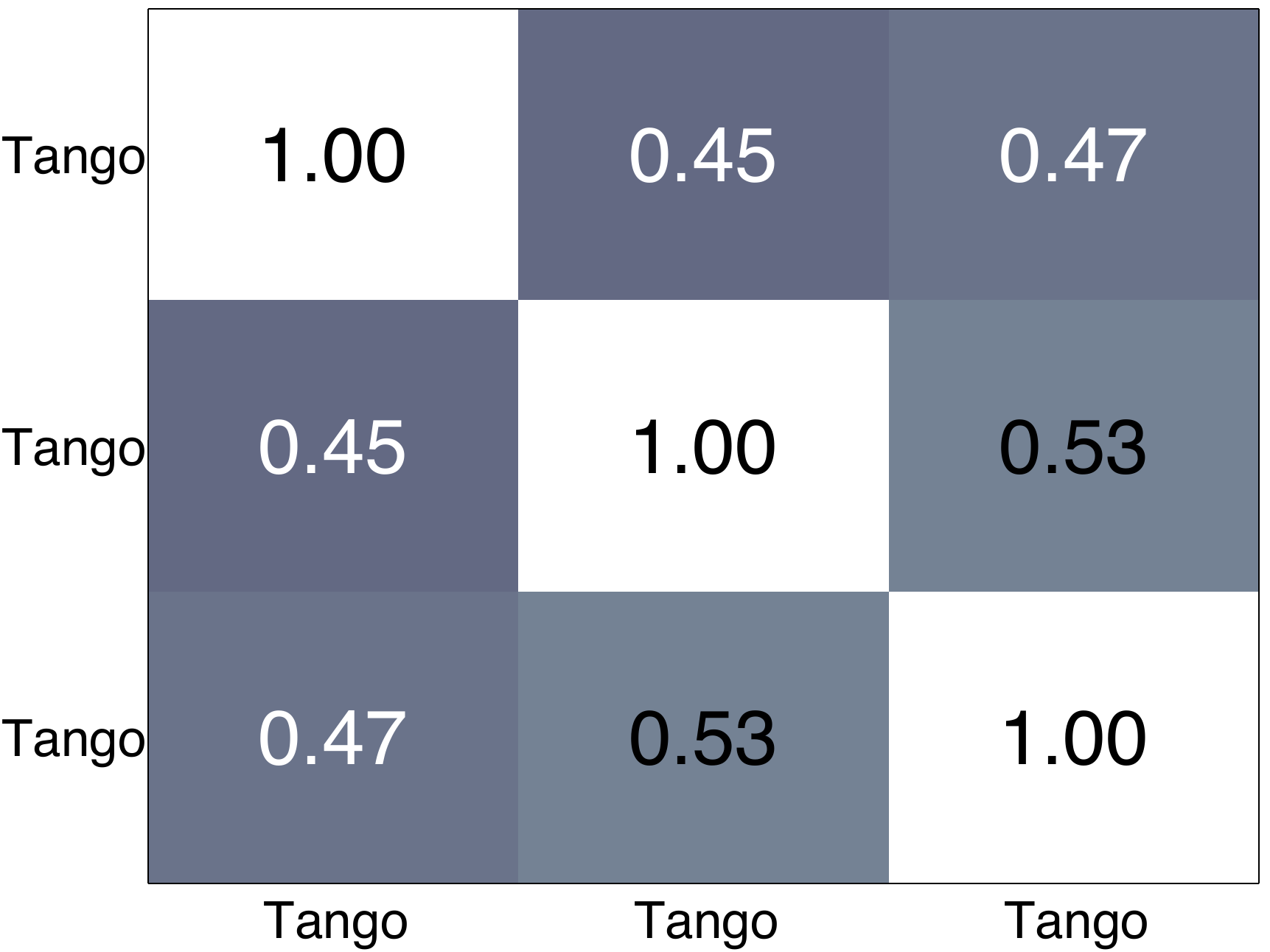} & 
    \includegraphics[width=.33\textwidth]{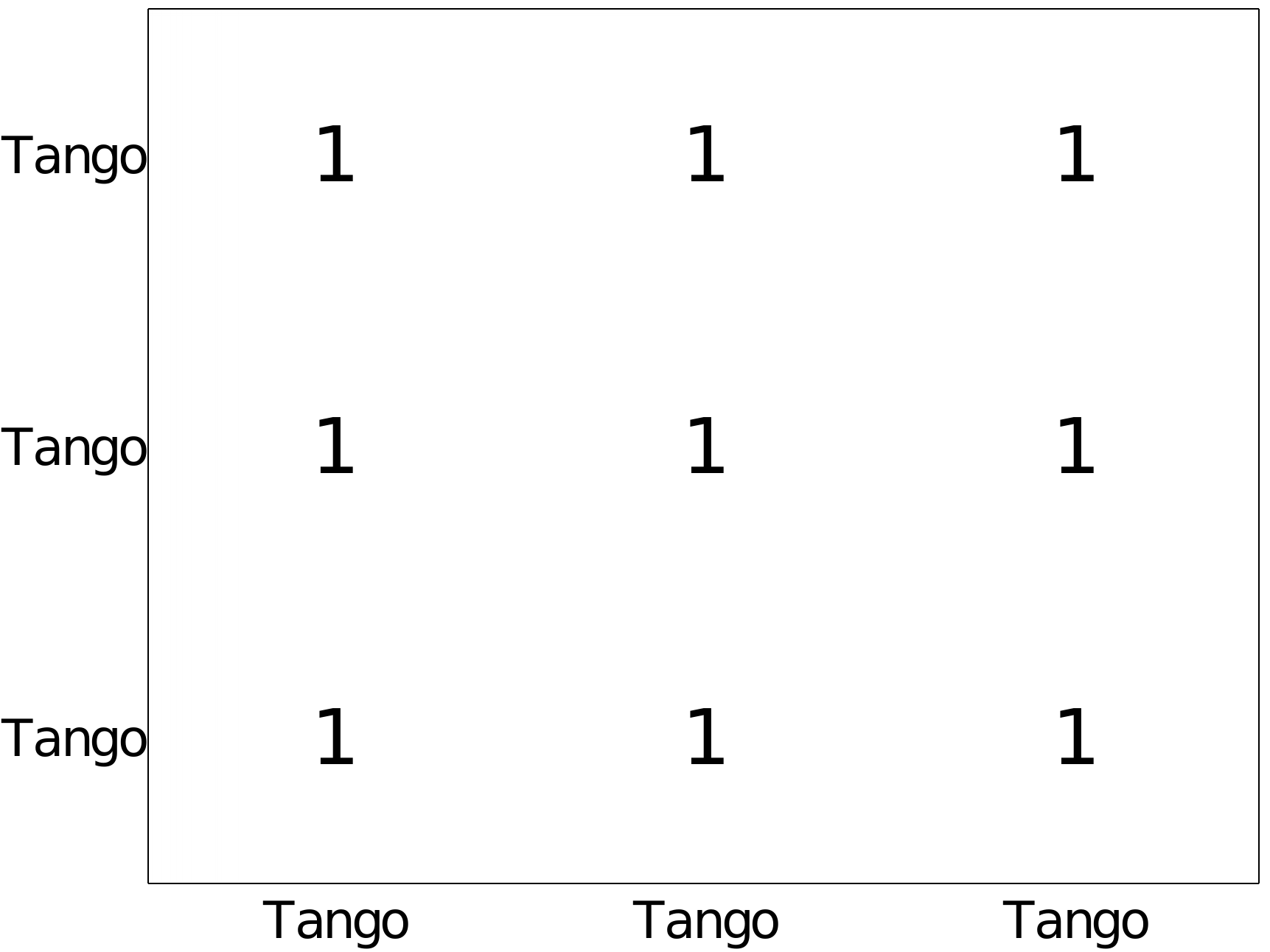} \\
    \end{tabular}
}

\subfloat[Two different activities are correctly identified (the couples on the left and right are grouped together, while the couple in the middle is isolated)]{
    \begin{tabular}{@{\hspace{0pt}}m{.27\textwidth}*{2}{@{\hspace{4pt}}m{.33\textwidth}}@{\hspace{0pt}}}
    \begin{minipage}{.27\textwidth}
    \centering
    \includegraphics[width=\textwidth]{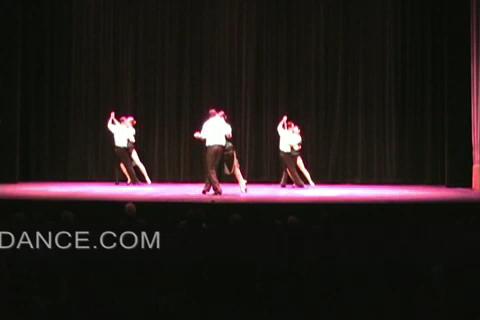}
    \includegraphics[width=\textwidth]{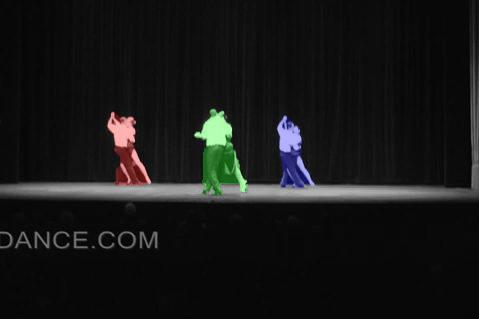} 
    \end{minipage} &
    \includegraphics[width=.33\textwidth]{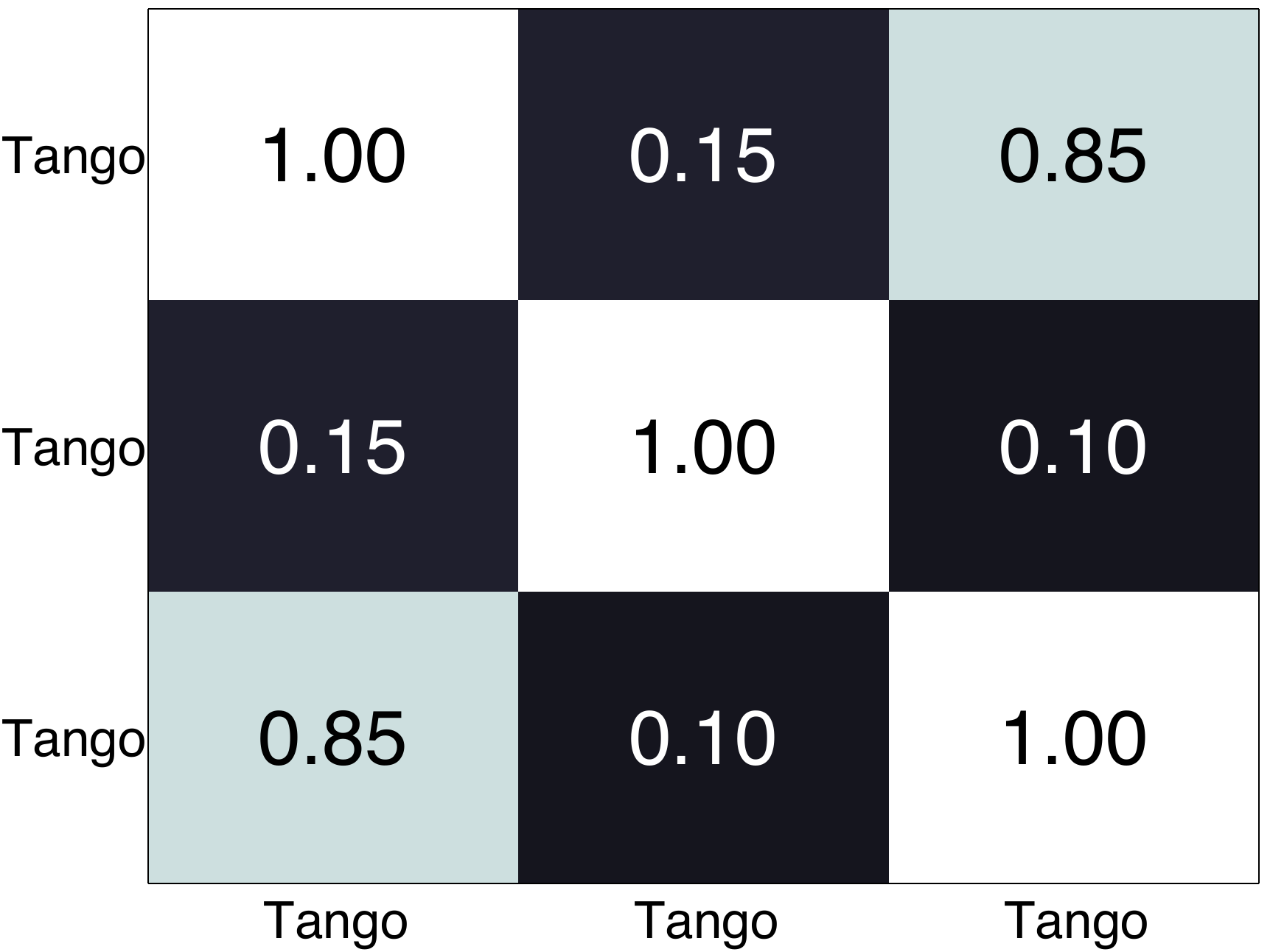} & 
    \includegraphics[width=.33\textwidth]{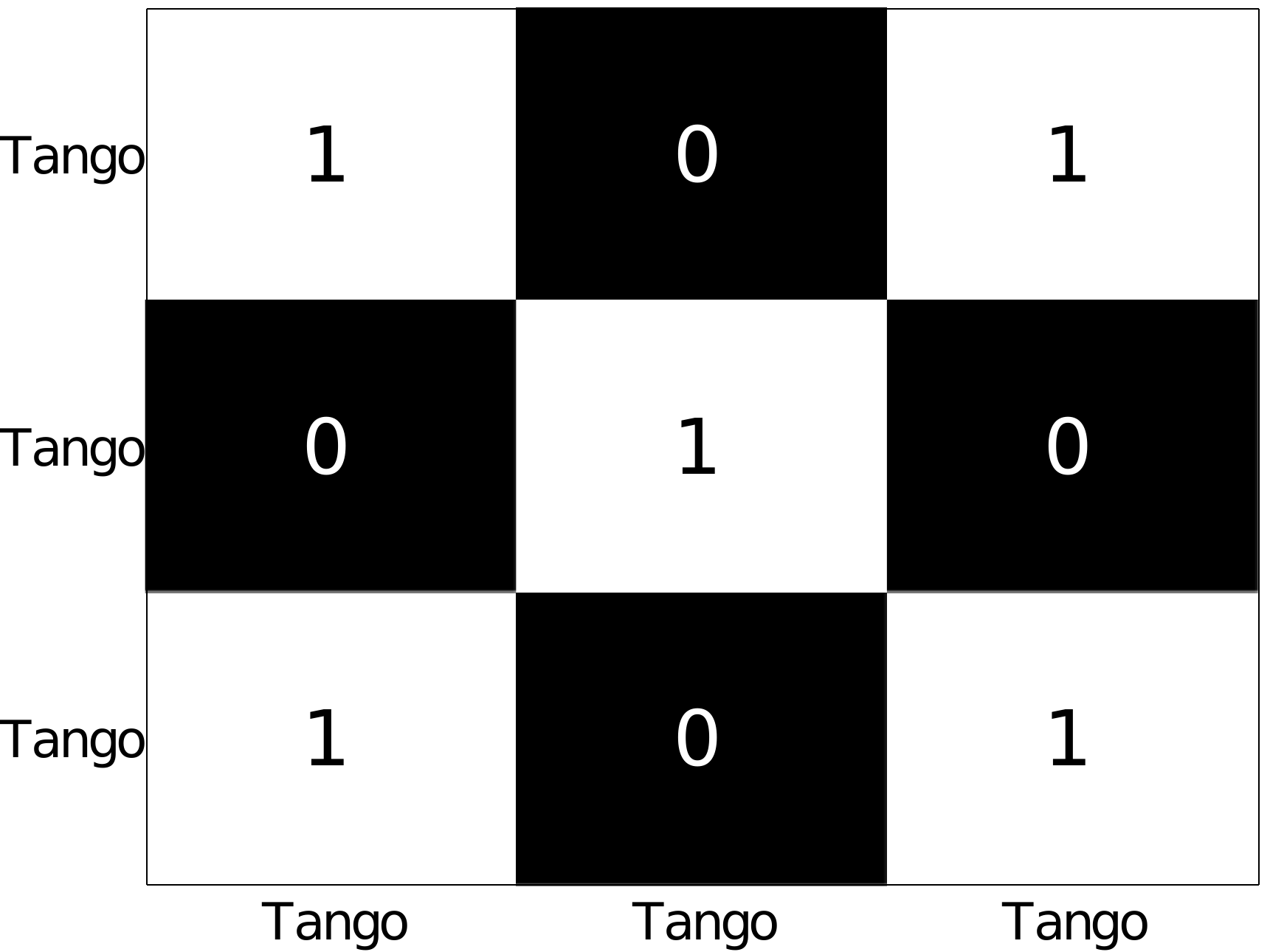} \\
    \end{tabular}
}

\caption{ \textbf{Left}: Sample frames from the Tango video, where three couples are dancing Tango during a one second interval were tracked/segmented (masks are displayed in different colors on the bottom left). Each couple was treated as a single entity. \textbf{Center}: affinity matrix. \textbf{Right}: Binarized affinity matrix after applying the threshold $\tau = 0.9/2 = 0.45$. (This is a color figure.)}
\label{fig:Tango_dancing_example}
\end{figure}

Two additional tests were performed using the Tango video, where there are three couples dancing Tango (Fig~\ref{fig:Tango_dancing_example}). Instead of treating each individual separately, we considered each couple as a single entity and applied the proposed method. The returned affinity matrix shows that the proposed method correctly groups the three couples as 
one group (if they are performing the same activity) or two groups (if they are performing different activities).\footnote{A video with the results of this experiments is available at: \url{http://youtu.be/WwAjSU_RuXA}. Bounding boxes of different color indicate different actions.} 

Let us now turn our attention to the employed features to point out the effectiveness of our simple approach.
We conducted experiments with 24 configurations of the video segments from the Long jump, the Gym, and the Kids videos, see figs.~\ref{fig:imgs_running_long_jumping},~\ref{fig:imgs_punching_dancing}, and~\ref{fig:imgs_kids} (8 configurations from each of them, similar to the configurations in tables~\ref{table:running_long_jumping} and \ref{table:punching_dancing}). We compared our simple feature (temporal gradient detector) against several feature detectors (the cuboid \citep{dollar05}, and Harris 2D \citep{harris_corner} (also see the local motion patterns (LMP) by \citet{Guha2012}) and descriptors (our 3D temporal gradient patches, HOG3D \citep{klaser08}, and the cuboid \citep{dollar05}). The Harris 2D detector detects spatially distinctive corners in each image frame, while the cuboid detector relies on applying separable linear filters, which produce high responses at local image intensities containing periodic frequency components. It also responds to any region with spatially distinct characteristics undergoing a complex motion \citep{dollar05}. It is important to mention that both detectors produce fewer feature points than the temporal gradient detector. As for the descriptors, they all produce vectors with comparable dimensionalities: $m=1,575$ for the temporal gradient patch, $m=1,440$ for the cuboid descriptor, and $m=1,000$ for HOG3D. The results are presented in Table~\ref{tab:compare_feature}, where the best grouping performance is obtained with the proposed detector/descriptor based on the temporal gradient. 
According to the evaluation by \citet{wang09}, HOG3D performs well in combination with dense sampling, which can capture some context information. But this is not appropriate in our unsupervised grouping framework for single videos. 
The cuboid detector gives good results in combination with the temporal gradient descriptor and HOG3D, but the cuboid descriptor seems to under perform.
The Harris 2D detector only produces 2D feature points, which do not necessarily undergo significant motion over time. 
Even though we employ temporal gradient here, we are not claiming that the other features are intrinsically or generally bad, since they work extremely well on supervised scenarios.  Nevertheless, for the problem at hand, where data is scarce, our simple feature performs better. 

\begin{table}[ht]

\caption{The grouping classification accuracy on 24 example configurations from the Long jump, the Gym, and the Kids videos by using several detectors/descriptor combinations.}
\label{tab:compare_feature}

\centering
\begin{threeparttable}[b]

\begin{tabular}{lccc} 
\toprule
\multicolumn{1}{c}{\multirow{2}{*}{Detector}} & \multicolumn{3}{c}{Descriptor} \\
\cmidrule{2-4}
& Temporal gradient & HOG3D\tnote{a} 
& Cuboid\tnote{b}  \\
\midrule
Temporal gradient   & \textbf{87.5\%} & 75.0\% & ---\tnote{d}  \\
Cuboid\tnote{b}      & 79.1\% & 79.1\% & 66.7\%  \\
Harris 2D\tnote{c}   & 75.0\% & 75.0\% & ---\tnote{d} \\
\bottomrule
\end{tabular}
\begin{footnotesize}
\begin{tablenotes}
\begin{scriptsize}
\item[a] \citet{klaser08}, implementation available at \url{http://lear.inrialpes.fr/~klaeser/software_3d_video_descriptor}.
\item[b] \citet{dollar05}, implementation available at \url{http://vision.ucsd.edu/~pdollar/files/code/cuboids/}.
\item[c] Harris points are detected in each frame. Patches around these keypoints are used to construct the spatio-temporal descriptors. This is similar to the local motion patterns (LMP) proposed in \citet{Guha2012}.
\item[d] A separate implementation of the cuboid descriptor is not available.
\end{scriptsize}
\end{tablenotes}
\end{footnotesize}

\end{threeparttable}

\end{table}

\subsection{Temporal analysis experiments}\label{sec:time_exp}
To test the proposed strategy for dealing with temporal action changes, we processed several video configurations with two
consecutive time intervals. The main goal is to identify the individuals who changed actions.

Table~\ref{table:time_evolution_race} summarizes the results by applying the method described in Section~\ref{sec:time} to the
Long jump video (Fig.~\ref{fig:imgs_running_long_jumping}).
Correct results were obtained when analyzing a video subset of the involved persons changing their actions (experiments 3 and 4). In the case where all the persons change action simultaneously or keep doing the same action (experiments 1 and 2), we observe incorrect results. The proposed framework for representing actions is not the source of this issue. It is a consequence of the normalization in Equation~(\ref{eq:temporalEnergyVector}), which compares individuals who are either all changing their action or all continuing their previous action and hence provides no discriminative power in this case.

\begin{figure}[ht]
\centering
\includegraphics[width=0.45\textwidth]{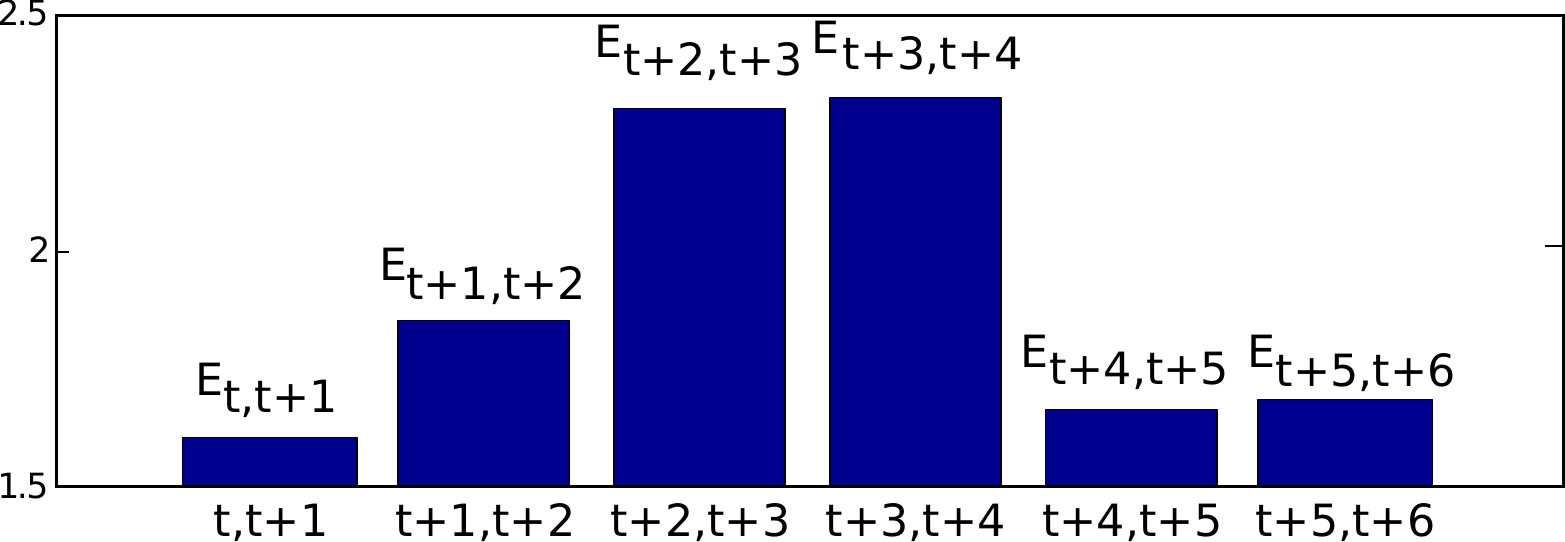}
\caption{The vector $ [\evol^{i}_{t,t+1}, \evol^{i}_{t+1,t+2}, ...
\evol^{i}_{t+5,t+6}] $ (see Equation~(\ref{eq:temporalEnergyVector}, p.~\pageref{eq:temporalEnergyVector})) over seven consecutive time intervals for one individual in the Gym video (see Fig.\ref{fig:warmingup_clustering}), during which he/she performs two different actions. We can see that the values of $\evol^{i}_{t,t+1}$, $\evol^{i}_{t+1,t+2}$,  $\evol^{i}_{t+4,t+5}$, and $\evol^{i}_{t+5,t+6}$ are small, reflecting no change of action in that interval. The other values ($\evol^{i}_{t+2,t+4}$ and $\evol^{i}_{t+3,t+4}$) are relatively big because the individual is changing actions. The transition between actions is not instantaneous, lasting for about two seconds.}
\label{fig:action_evolution}
\end{figure}

Although it is not easy to extract a general rule for every possible scenario, the vector $ [\evol^{i}_{t-1,t}, \evol^{i}_{t,t+1},
\evol^{i}_{t+1,t+2}, ...] $ (see Equation~(\ref{eq:temporalEnergyVector}, p.~\pageref{eq:temporalEnergyVector})) provides useful information about how an individual's actions evolves over time.
An example is shown in Fig.~\ref{fig:action_evolution}, where we build this vector for one individual in the Gym video (see Fig.\ref{fig:warmingup_clustering}), using seven consecutive one-second time intervals (from $t$ to $t+6$). During this time lapse, the individual changes his action one time.
More complex rules can be derived from this readily available information. The action-change rule provided in Section~\ref{sec:time} will nonetheless be already useful in many cases. 


\begin{table}[ht]

\caption{Temporal analysis of the Long jump video. Three persons in a race track on consecutive time intervals.
`R' and `LJ' denote running and long-jumping, respectively. A value
above $\mu=0.3$  in the action evolution vector $\Evol_{t-1,t}$ means
that person's action has changed.}
\label{table:time_evolution_race}

\centering
\begin{tabular*}{\textwidth}{@{\extracolsep{\fill}} *{13}{c}}
\toprule

\multirow{3}{*}{Person} &
\multicolumn{3}{c}{Experiment 1} &
\multicolumn{3}{c}{Experiment 2} &
\multicolumn{3}{c}{Experiment 3} &
\multicolumn{3}{c}{Experiment 4} \\

&
\multicolumn{2}{c}{Interval} & \multirow{2}{*}{$\Evol_{t-1,t}$} &
\multicolumn{2}{c}{Interval} & \multirow{2}{*}{$\Evol_{t-1,t}$} &
\multicolumn{2}{c}{Interval} & \multirow{2}{*}{$\Evol_{t-1,t}$} &
\multicolumn{2}{c}{Interval} & \multirow{2}{*}{$\Evol_{t-1,t}$} \\

\cmidrule{2-3}
\cmidrule{5-6}
\cmidrule{8-9}
\cmidrule{11-12}

&
$t-1$ & $t$ & &
$t-1$ & $t$ &  &
$t-1$ & $t$ &  &
$t-1$ & $t$ &  \\

\midrule

A &
\actionA{R} & \actionA{R} & \textbf{0.378} &
\actionA{R} & \actionB{LJ} & \textbf{0.309} &
\actionA{R} & \actionA{R} & 0.133 &
\actionA{R} & \actionA{R} & 0.202 \\

B &
\actionA{R} & \actionA{R} & \textbf{0.336} &
\actionA{R} & \actionB{LJ} & \textbf{0.353} &
\actionA{R} & \actionB{LJ} & \textbf{0.437} &
\actionA{R} & \actionA{R} & 0.142 \\

C &
\actionA{R} & \actionA{R} & 0.286 &
\actionA{R} & \actionB{LJ} & \textbf{0.338} &
\actionA{R} & \actionB{LJ} & \textbf{0.430} &
\actionA{R} & \actionB{LJ} & \textbf{0.656} \\
\bottomrule
\end{tabular*}

\end{table}

\begin{figure}[ht]
 \centering
 \includegraphics[width=0.23\textwidth,height=0.1\textheight]{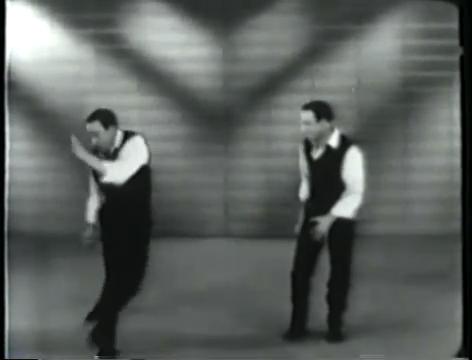}
 \includegraphics[width=0.23\textwidth,height=0.1\textheight]{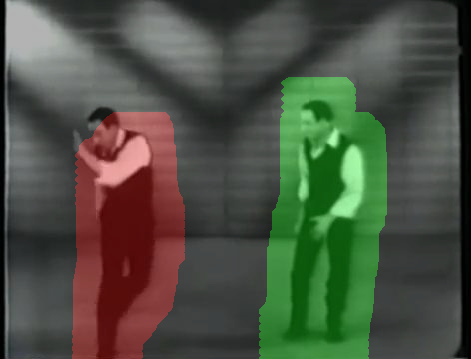}
 \includegraphics[width=0.23\textwidth,height=0.1\textheight]{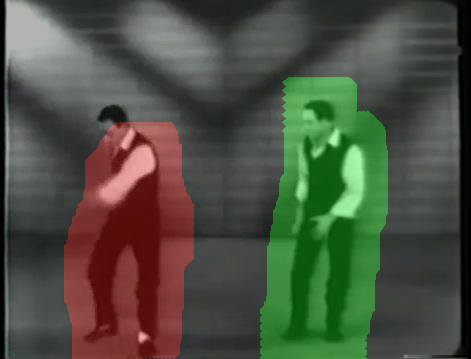}
 \includegraphics[width=0.23\textwidth,height=0.1\textheight]{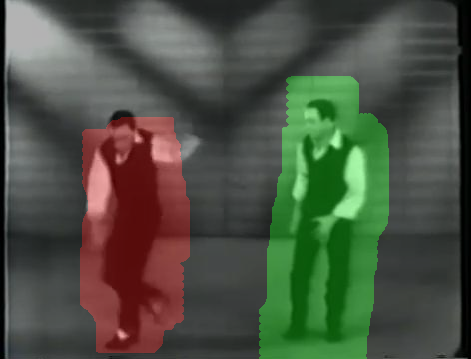}
\\
 \vspace{2pt}
 \includegraphics[width=0.23\textwidth,height=0.1\textheight]{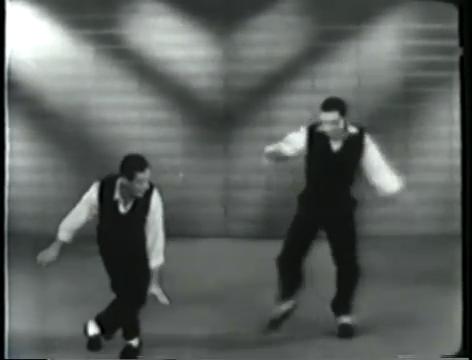}
 \includegraphics[width=0.23\textwidth,height=0.1\textheight]{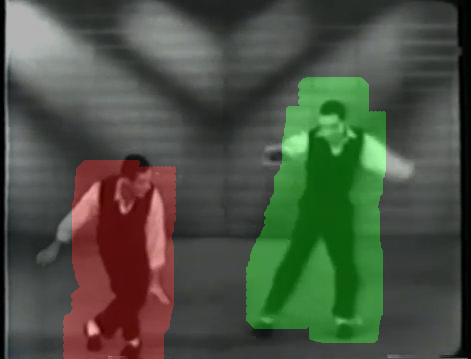}
 \includegraphics[width=0.23\textwidth,height=0.1\textheight]{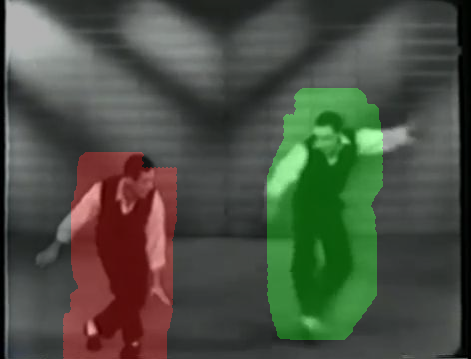}
 \includegraphics[width=0.23\textwidth,height=0.1\textheight]{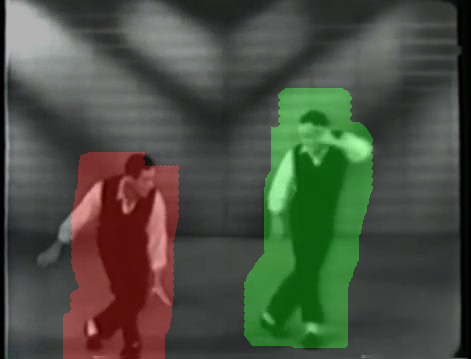}
 \caption{On the first column, two frames from the Mimicking video.
  On the remaining columns, the tracking/segmentation masks are
displayed in colors.
  The two dancers are correctly detected as performing different actions.
  (This is a color figure.)}
 \label{fig:mimicking}
\end{figure}

Finally, we present an example using the Mimicking video for the special case (P=2) described in
Section~\ref{sec:special}. It consists of two seconds interval ($t-1$ and $t$)
from a comedy show, where two dancers ($i$ and $j$) are mimicking each
other (see  Fig.~\ref{fig:mimicking}).  Using
equations~(\ref{eq:space_special_case}) and
(\ref{eq:time_special_case}), and a threshold $\mu = 1/2$, we obtain
$\evol^{i,j}_{t-1} = 2.29$,  $\evol^{i,j}_t = 1.07$,  $\evol^{i}_{t-1,
t} = 1.62$, and $\evol^{j}_{t-1,t} = 1.64$. These results correctly imply that
the dancers were performing different actions on each of the two
seconds, and that both dancers went through an action change from
$t-1$ to $t$.\footnote{Videos with the results of this experiment is available at: \url{http://youtu.be/I922vARiGko}, \url{http://youtu.be/skfJSs5-ftI} and \url{http://youtu.be/SO4x8YeSbao}. Bounding boxes of different color indicate different actions.}


\subsection{Joint spatio-temporal grouping} \label{sec:spatio_temporal_grouping}
We further analyzed three videos, i.e., the Long jump, Fitness, and Outdoor videos, in which human actions are jointly grouped in time and space (recall that by space we mean across individuals), applying the nonparametric spectral clustering algorithm by \citet{Perona2004} to the pairwise affinity matrix described in Section~\ref{sec:joint_grouping}.

\begin{figure}[ht]
 \centering
 \includegraphics[width=1.0\textwidth]{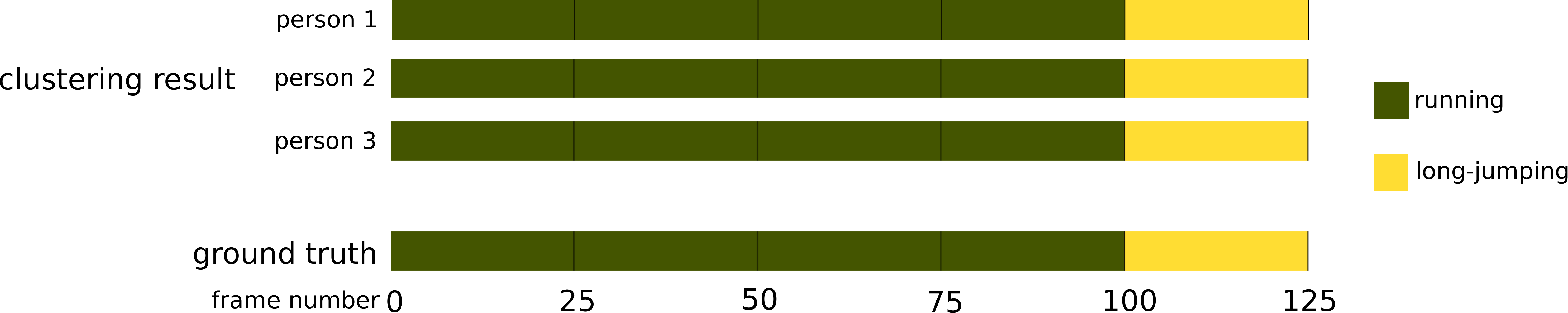} 
 \caption{5 seconds from the Long jump video, where three athletes are running and long-jumping, see sample frames in Fig.~\ref{fig:imgs_running_long_jumping}. Our method correctly identifies two actions.}
 \label{fig:long_jumping_clustering}
\end{figure}

The first experiment, using $5$ seconds from the Long jump video is shown in Fig.~\ref{fig:long_jumping_clustering}. The $3$ individuals are first running then long-jumping. We thus consider that there are $15 = 5 \cdot 3$ individuals in the video.
The clustering algorithm on this $15 \times 15$ affinity matrix gives $2$ correct clusters, even though the three athletes have different appearance.\footnote{A video with the results of this experiment is available at: \url{http://youtu.be/9KQa9mFXBIk}.}

\begin{figure}[ht]
 \centering
 \includegraphics[width=0.23\textwidth]{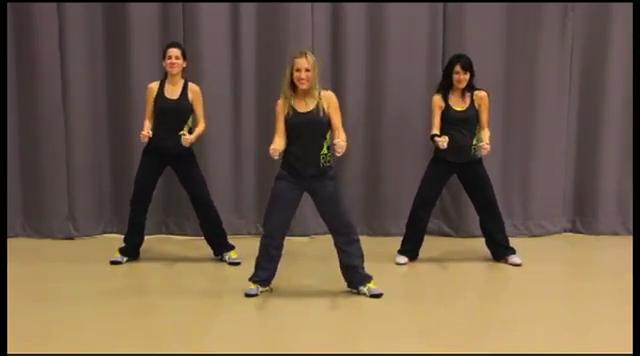} 
 \includegraphics[width=0.23\textwidth]{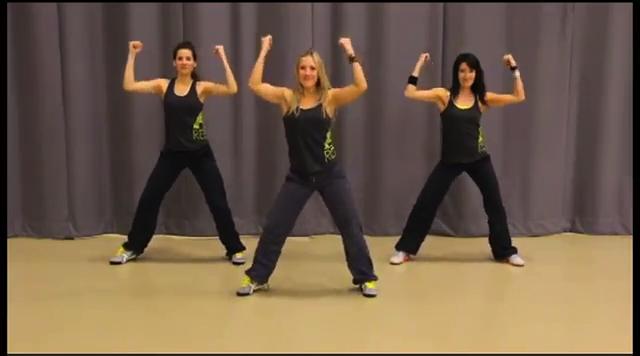} 
 \includegraphics[width=0.23\textwidth]{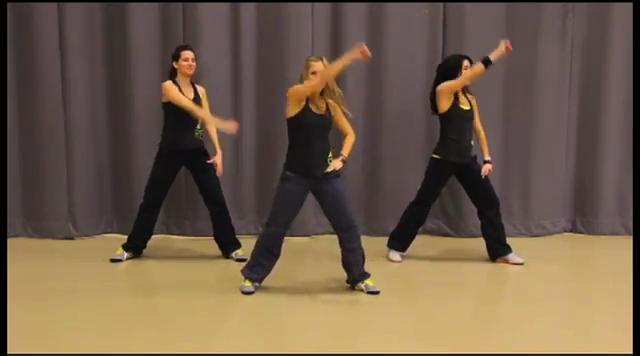} 
 \includegraphics[width=0.23\textwidth]{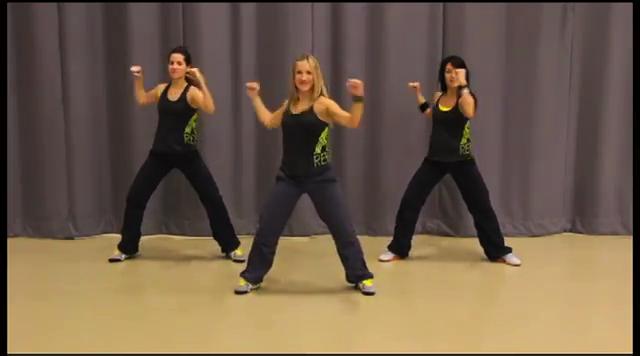} \\
 \vspace{4ex}
 \includegraphics[width=1.0\textwidth]{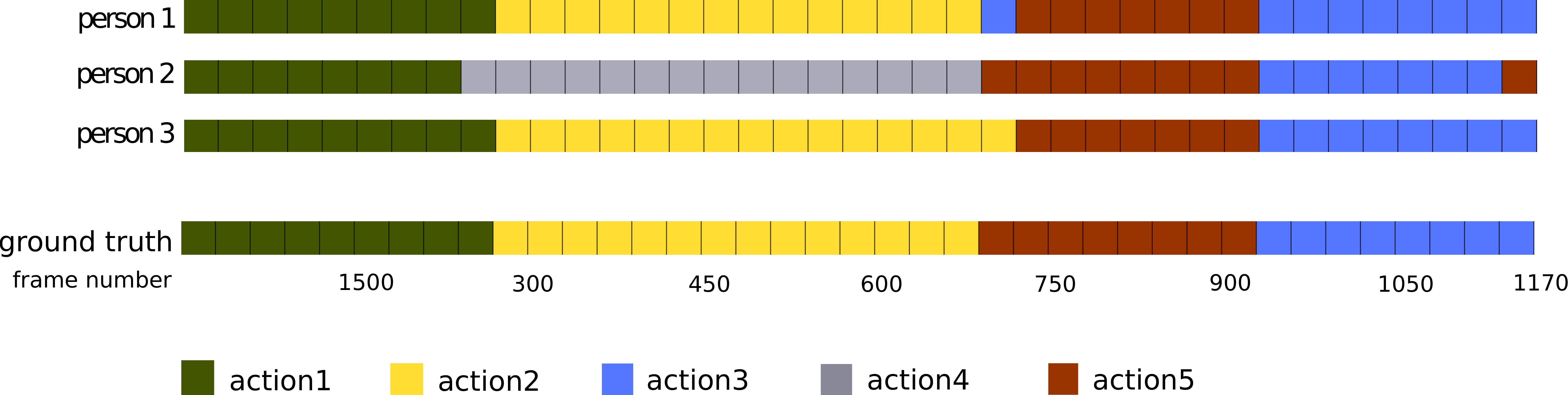} 
 \caption{40 seconds analysis of the Fitness video, where five clusters are detected instead of four. Between frames 300 and 690, all three persons are doing the same corase action and this over-splitting can be explained by granular action variability, where the person in the middle presents auto-similarity (she is somewhat more energetic than the others).}
 \label{fig:warmingup_clustering}
\end{figure}

The second experiment was conducted on a 40 seconds sequence from the Fitness video, in which three individuals are doing gym exercises (Fig.~\ref{fig:warmingup_clustering}). Notice that, even though their actions are synchronized, we do not provide this information {\it a priori} to the algorithm. The clustering algorithm returned $5$ clusters from the $120 \times 120$ affinity matrix, and $4$ should have been found.\footnote{A video with the results of this experiment is available at: \url{http://youtu.be/5moGG2e4PXc}.} There is an over splitting for the individual in the middle from frame 300 to frame 690, meaning that in this case either auto-similarity was captured in excess  or the action of this person is actually different (such granularity in the action can actually be observed by carefully watching the video). 
There are also some clustering incorrect results in the transition period between two actions due to temporal mixing effects (our intervals are arbitrarily fixed and may incorrectly divide the actions during the switch). Note also that the ground truth was manually built from visual observation and is also fixed to having hard action transitions. Considering overlapping or shorter segments will alleviate this issue if additional temporal accuracy is needed.

\begin{figure}[ht]
 \centering
 \begin{tabular}{@{\hspace{0pt}}*{3}{c@{\hspace{4pt}}}c@{\hspace{0pt}}}
 \includegraphics[width=0.235\textwidth]{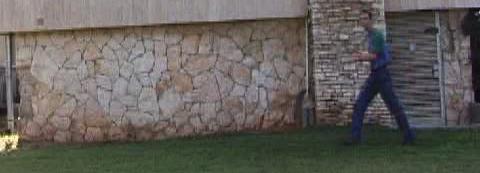} &
 \includegraphics[width=0.235\textwidth]{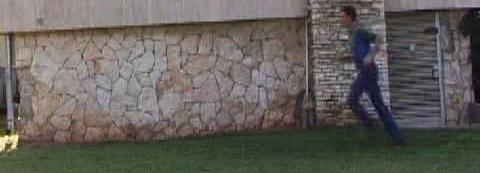} &
 \includegraphics[width=0.235\textwidth]{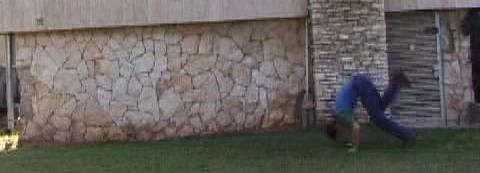} &
 \includegraphics[width=0.235\textwidth]{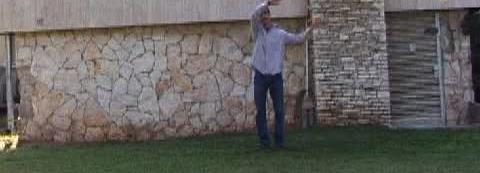} \\
 \includegraphics[width=0.235\textwidth]{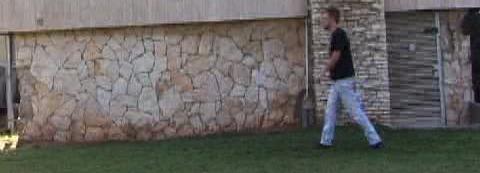} &
 \includegraphics[width=0.235\textwidth]{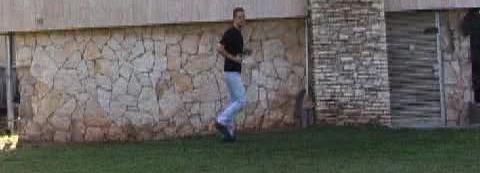} &
 \includegraphics[width=0.235\textwidth]{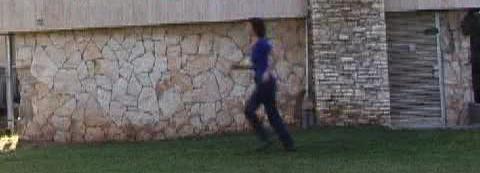} &
 \includegraphics[width=0.235\textwidth]{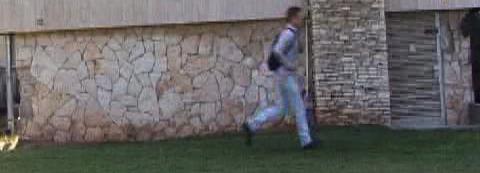} \\
 \end{tabular}
 \vspace{4ex}
 
 \includegraphics[width=1.0\textwidth]{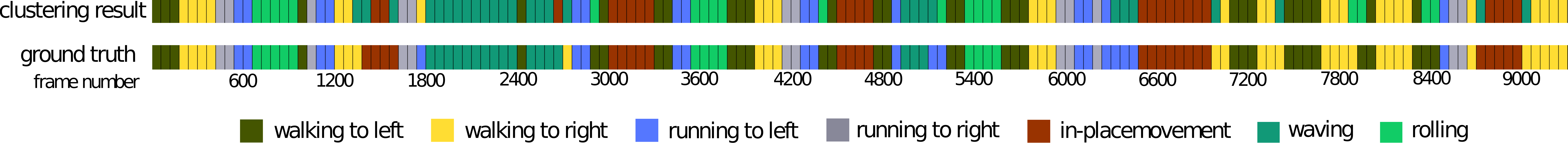} 
 
 \caption{320 seconds from the Outdoor video. In accordance with visual observation, seven clusters are identified. There are a few clustering errors in the transition periods between the actions due to the discrete temporal nature of the particular analysis here exemplified.}
 \label{fig:outdoor_clustering}
\end{figure}

In the third experiment we processed the Outdoor video (Fig.~\ref{fig:outdoor_clustering}). The video contains only one individual per frame, which changes appearance (different individuals appear over time with different clothing). This video exhibits very slow motions and in order to capture enough action information in the temporal direction, we first subsampled the 
video by a factor of $2$ in the temporal direction before applying the proposed method. The clustering is consistent with visual observation.\footnote{A video with the results of this experiment is available at: \url{http://youtu.be/1-XO-D9qRRg}.} We observe some incorrect labels, again due to the fixed transition period between two 
actions, that is, one time interval can contain two actions and the action type is not well defined in this situation for the corresponding time segment.

\section{Concluding remarks}\label{sec:conclusion}
We presented an unsupervised sparse modeling framework for action-based scene analysis from a single video. We model each of the individual actions independently, via sparse modeling techniques, and build a group affinity matrix. Applying relatively simple rules based on representation changes, the proposed method can efficiently and accurately tell whether there are different actions occurring on the same short-time interval,  and across different intervals, including detecting possible action changes by any of group members. In addition, we extended the method to handle longer motion imagery sequences by applying standard spectral clustering techniques to a larger spatio-temporal affinity matrix. 
 We tested the performance of the framework with diverse publicly available datasets, demonstrating its potential effectiveness for diverse applications.

We also showed that by using a single and simple feature in such a scarce data scenario outperforms standard and more sophisticated ones.
This indicates that further research on good features for unsupervised action classification is much needed.

We are currently working on extending the model to handle interactions, that is, meta-actions performed by several persons. Also, going from purely local features to semi-local ones by modeling their interdependences might provide a way to capture more complex action dynamics. Finally, action similarity is an intrinsically multiscale issue (see example in Figure 14, where the middle lady is performing the same coarse action but in a different fashion), therefore calling for the incorporation of such concepts in action clustering and detection. 

%
%

%

\section*{Acknowledgements}
We thank N.~Papadakis for kindly providing the tracking/segmentation software used in all the experiments.


\bibliographystyle{spbasic}
\bibliography{references.bib}
\end{document}